\documentclass[sigconf,nonacm]{acmart}
\renewcommand\footnotetextcopyrightpermission[1]{}
\AtBeginDocument{%
  }

\usepackage{xcolor}
\usepackage{multirow}     
\usepackage{soul}

\begin{document}

\title{
  Partial Knowledge Distillation for Alleviating the Inherent Inter-Class Discrepancy in Federated Learning
}


\author{Xiaoyu Gan$^1$, Jingbo Jiang$^2$, Jingyang Zhu$^3$, Xiaomeng Wang$^2$, Xizi Chen$^1$, Chi-Ying Tsui$^2$}

\affiliation{$^1~$College of Informatics \country{} Huazhong Agricultural University}
\affiliation{$^2~$Department of Electronic and Computer Engineering \country{} The Hong Kong University of Science and Technology}
\affiliation{$^3~$NVIDIA Corporation \country{}}

%%%%%%%%%%% author template %%%%%%%%%
% \author{name}
% \affiliation{%
%   \department{xx department}
%   \institution{xx institution}
%   \city{xx city}
%   \country{xx country}
% }
% \email{email@xx.xx}
% \authornote{Corresponding author.}


\begin{abstract}
  Substantial efforts have been devoted to alleviating the impact of the long-tailed class distribution in federated learning. In this work, we observe an interesting phenomenon that certain weak classes consistently exist even for class-balanced learning. 
  These weak classes, different from the minority classes in the previous works, are inherent to data and remain fairly consistent for various network structures, learning paradigms, and data partitioning methods. The inherent inter-class accuracy discrepancy can reach over 36.9\% for federated learning on the FashionMNIST and CIFAR-10 datasets, even when the class distribution is balanced both globally and locally.
  In this study, we empirically analyze the potential reason for this phenomenon. Furthermore, a partial knowledge distillation (PKD) method is proposed to improve the model's classification accuracy for weak classes. In this approach, knowledge transfer is initiated upon the occurrence of specific misclassifications within certain weak classes.
  Experimental results show that the accuracy of weak classes can be improved by 10.7\%, reducing the inherent inter-class discrepancy effectively.
\end{abstract}



\maketitle

\section{Introduction}
\label{sec: 1}
\begin{figure}[!t]
  \centering
  \setlength{\belowcaptionskip}{-3pt}
  \includegraphics[width=0.49\textwidth]{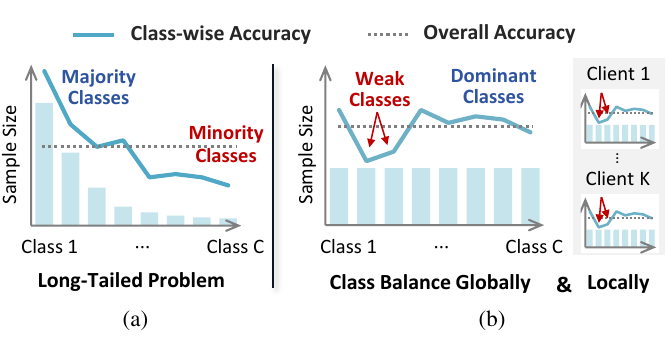}
  \caption{
    (a) The Conventional Long-tailed Problem; 
    (b) Inherent Inter-class Discrepancies Observed under Balanced Class Distributions (both Globally and Locally).
    }\label{fig: 1}
\end{figure}

Federated learning (FL) is a decentralized machine learning paradigm where multiple clients collaborate to train a global model without directly sharing their local data \cite{fedavg}. This approach preserves data privacy by retaining the data on local devices. However, a big challenge to FL is the varying performance across different classes due to the presence of unbalanced sample sizes \cite{surveyCIFL}. When the number of samples for different classes varies substantially, the global model often exhibits bias towards the majority classes, known as `head classes'. The minority classes, referred as `tail classes', become underrepresented \cite{surveyLT}, thus leading to lower classification accuracy as depicted in Fig.~\ref{fig: 1}(a). However, in this work, we show that `dominant' and `weak' classes consistently exist even when sample sizes are equalized and samples from each class are uniformly distributed across the clients (i.e., with a balanced class distribution both globally and locally).
Unlike the minority classes discussed in prior studies, these weak classes are intrinsic to the data itself, rather than being dictated by sample size. The resulting inter-class accuracy discrepancy (ICD) is illustrated in Fig.~\ref{fig: 1}(b).

\begin{figure*}[!t]
  \centering
  \setlength{\belowcaptionskip}{8pt}
  \includegraphics[width=0.97\textwidth]{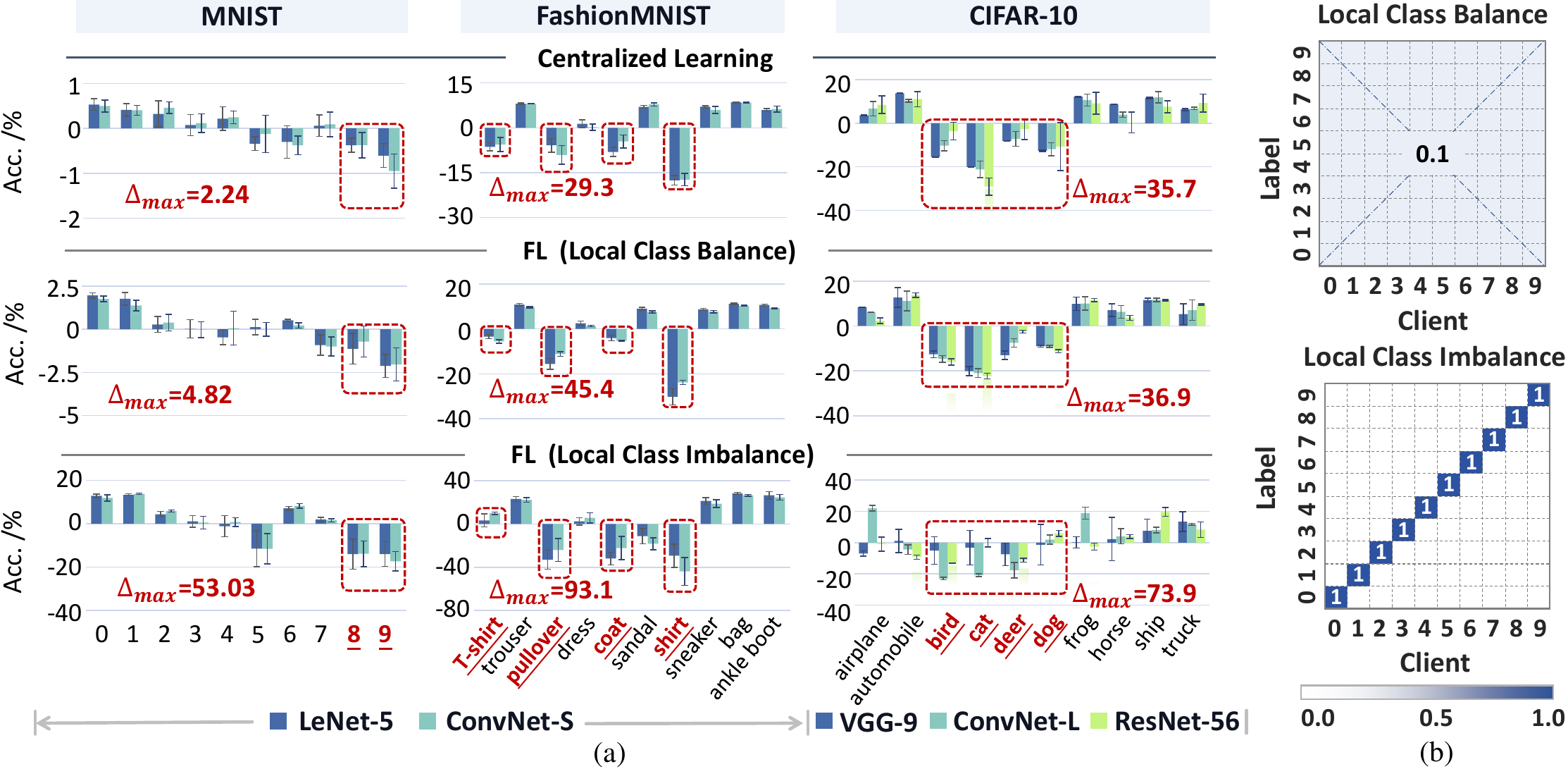} 	
  \caption{
    (a) The Class-wise Accuracy (with Mean Subtraction) Based on Different Learning Paradigms and Network Structures ($\Delta_{max}$ Refers to the Maximum Discrepancy after a Training Session; the Network Structures, such as VGG-9 \cite{vgg,FedMA} and ConvNets \cite{fedavg}, are Detailed in Section~\ref{sec: 4.1}.)     
    (b) Sample Distribution in Different Scenarios (Assuming a Total of 10 Clients).
  }\label{fig: 2}
\end{figure*}

Fig.~\ref{fig: 2} presents the class-wise accuracy across several widely used benchmarks in the domain of FL. Each experiment is repeated five times, and the average result is presented. The ICD phenomenon identified in this study exhibits the following characteristics. 
1)~This phenomenon is universally present across different \textit{learning paradigms} and is particularly pronounced in FL compared to centralized learning.  
2) Neural networks trained on the same dataset tend to exhibit a very similar ICD, indicating that this phenomenon is irrelevant to the \textit{network structure}.
3) This phenomenon persists no matter whether the \textit{local class distributions} in FL is balanced or not. Notably, as local imbalance intensifies, particularly when samples from weak classes are concentrated within a limited number of clients, the inherent ICD becomes significantly exacerbated. Details will be elaborated upon in the subsequent sections. 

%% Figure 1: 

%% Figure 2: 

Although promising progress has been made in alleviating the inter-class discrepancy brought by class imbalance \cite{surveyLT}, these techniques cannot be directly applied to the inherent ICD problem. Specifically, current approaches usually focus on rebalancing class distributions by employing techniques such as over-sampling \cite{resample} or data augmentation \cite{LEAP} for the minority classes. However, considering that the inherent ICD is not associated with the quantity of samples, these methods may not be sufficient to mitigate the problem effectively. 

Experiments indicate that a likely reason for this ICD phenomenon is the intrinsic similarities in high-level features between specific classes.
This similarity poses a challenge for the classifier in accurately distinguishing data samples across these classes.
In this work, we propose a \textbf{federated partial knowledge distillation} (PKD) method to improve classification accuracy for weak classes. Our method is inspired by the observation that an expert trained specifically on a group of confounding weak classes show greater proficiency in differentiating among them. This specialized model focuses solely on extracting and distinguishing the subtle feature differences among these classes and thus yields better accuracy than the model trained jointly on all classes. The term `\textbf{partial}' in PKD is shown in several aspects:  1) An expert's knowledge is confined to specific classes. It is tailored to identify a subset of data that shares analogous high-level features. 2) Knowledge distillation (KD) is triggered only when misclassifications between certain classes happens. Therefore, only a subset of the misclassified samples from these weak classes is considered in computing the KD loss during the training phase.  3) While there might be multiple experts, only up to one expert can be involved in the KD process for any given misclassified sample. Experimental results demonstrate that the proposed PKD method can effectively alleviate the inherent ICD issue by enhancing the classification accuracy of the worst performing classes. In summary, this work makes the following contributions:

\begin{itemize}
\item We highlight an interesting phenomenon that certain weak classes are always present, even in ideal scenarios where class distributions are balanced at both the global and local levels. This inherent variance in accuracy across different classes is especially pronounced in the context of FL, reaching up to 45.4\% on the FashionMNIST dataset \cite{Fashion} and 36.9\% on the CIFAR-10 dataset \cite{Cifar10}, respectively. 
\item A federated partial KD method is proposed to enhance the learning process for weak classes. This approach selectively transfers knowledge from class-specific experts to the global student model, all within the FL framework.
\item Experimental results show that the proposed method can effectively improve the classification accuracy of the weak classes by over 10.7\% and 6.3\% on FashionMNIST and CIFAR-10, respectively.
\end{itemize}

\section{Inter-Class Accuracy Discrepancy}
\subsection{Discrepancy Induced by the Class Imbalance Problem}
\label{sec: 2.1}

\textbf{ICD due to class imbalance.}
A primary reason for the variance in class-wise accuracy in deep learning is the imbalanced class distribution \cite{imbalance_survey,imbalance_cnn}. The unbalanced sample sizes lead to a dual-fold challenge: 1) Models tend to be biased towards the majority classes, leaving the minority classes under-represented.  2) The lack of data for minority classes makes it even harder to learn the characteristics of these classes \cite{surveyLT}.
In federated learning, the potential causes for ICD become more complex. The presence of class imbalance at either the global level or within the local dataset of individual clients may diminish the model’s quality. Besides, the mismatch in class distributions between the cloud and the clients may also degrade the performance of the global model \cite{surveyCIFL}. 

\textbf{Existing approaches.} Existing solutions can be broadly classified into several categories. 
1) \textit{Class re-balancing techniques}: These methods focus on offsetting the negative impact of uneven class distribution in training samples. This can be achieved by either over-sampling the minority classes to ensure a more equitable presence in the training data \cite{resample,EMANATE}, or by re-weighting the training loss values to mitigate the influence of uneven positive gradients \cite{Focal,CBloss}.  
2) \textit{Data augmentation techniques}: These methods seek to enlarge the sample size or enhance the sample quality of minority classes through data augmentation techniques \cite{SMOTE, BalanceFL, m2m}. Compared to mere re-balancing methods, this approach can potentially enrich the diversity within the minority classes, thereby reducing the risk of overfitting. 
3) Other methods include meta-learning \cite{meta2}, ensemble learning \cite{LFME,ensemble}, transfer learning \cite{LFME}, and so on.
Although significant progress has been achieved in reducing the ICD induced by class imbalance, such strategies may not be directly applicable to addressing the inherent ICD problem, as we will discuss below.

\subsection{The Inherent ICD in This Work}
\label{sec: 2.2}
\textbf{Differences from the class imbalance problem.} While the weak classes show similar poor accuracy to the minority classes in the class imbalance problem, their underlying causes and potential solutions are distinct. First, the inherent ICD is not a result of uneven sample sizes across different classes. Weak classes consistently exist even in the ideal case where both global and local data are class balanced. Consequently, previous methods that rely on re-sampling local data \cite{BalanceFL} or clients \cite{clientsampl} based on their sample sizes are not feasible in this context.
Second, in contrast to minority classes, weak classes are not characterized by data scarcity or a lack of diversity. Experimental results suggest that the conventional data augmentation techniques \cite{cutout} fail to address the inherent ICD effectively, and in some cases, they might even intensify the inherent accuracy discrepancy.

\textbf{The inherent ICD phenomenon.}  Fig.~\ref{fig: 2}(a) shows the class-wise accuracy achieved on three datasets. In each scenario, the total number of samples for each class is kept identical, ensuring that global class balance is maintained. We adopt several commonly used convolutional neural networks and train the models following different learning paradigms. Two local data partitioning strategies are adopted in FL for each dataset, as illustrated in Fig.~\ref{fig: 2}(b). Our results indicate that this inherent ICD is a common phenomenon across various network structures, learning paradigms, and local data partitioning strategies. Most of the dominant classes and the weak classes, e.g., \{`0', `1'\} and \{`8', `9'\} in MNIST \cite{MNIST}, remain fairly consistent across various settings. The inter-class accuracy discrepancy is less marked for centralized learning and thus may go unnoticed. However, when it comes to FL, the discrepancy may increase significantly, e.g., reaching up to 45.4\% for FashionMNIST.
We also test the local class imbalance scenario using a pathological data partitioning strategy proposed in \cite{fedavg}. Specifically, each client is assigned samples from a single class to ensure a clear and straightforward comparison of the class-wise accuracy. As shown in Fig.~\ref{fig: 2}(a), the same group of weak classes is observed, with the accuracy discrepancy further increasing to a significant 53.0\% on the MNIST dataset. For more complex datasets like FashionMNIST and CIFAR-10, such discrepancy can reach up to over 73.9\%. 
When evaluated under FL, CIFAR-10 exhibits a reduced discrepancy ($\Delta$) compared to FashionMNIST.
This can be attributed to the degradation in overall model performance. For instance, under the extreme local class imbalance scenario, the average accuracy of VGG-9 \cite{vgg, FedMA} and ResNet-56 \cite{ResNet} notably drops below 20\% on CIFAR-10.

\begin{figure}[!t]
  \centering
  \setlength{\belowcaptionskip}{-11pt}
    \includegraphics[width=0.487\textwidth]{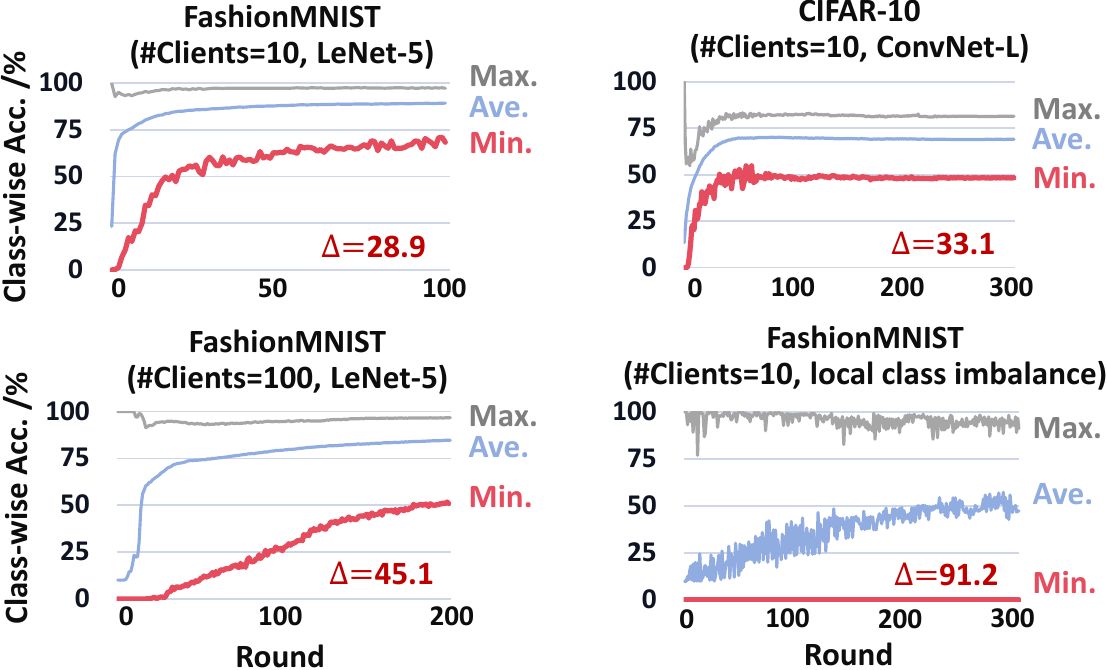} 	
    \caption{The Maximum, Average, and Minimum Class-wise Accuracy During Training. 
    }
    \label{fig: 3}
\end{figure}

Fig.~\ref{fig: 3} shows the maximum and minimum class-wise accuracy at each round. While the overall accuracy rises smoothly over time, the difference between the maximum and minimum values continues to be evident. With a balanced class distribution globally and locally, the accuracy discrepancy on the FashionMNIST dataset is 28.9\% when there are 10 clients.
This value increases significantly as the number of clients grows to 100. In scenarios with local class imbalance, samples from weak classes are concentrated within a small subset of clients, thereby increasing the accuracy discrepancy to 91.2\%. The minimum class-wise accuracy is close to zero, indicating that the final model might completely fail in specific classes.

%% Figure 3: 

\begin{figure*}[!t]
  \centering
  \setlength{\belowcaptionskip}{14pt}
    \includegraphics[width=1.0\textwidth]{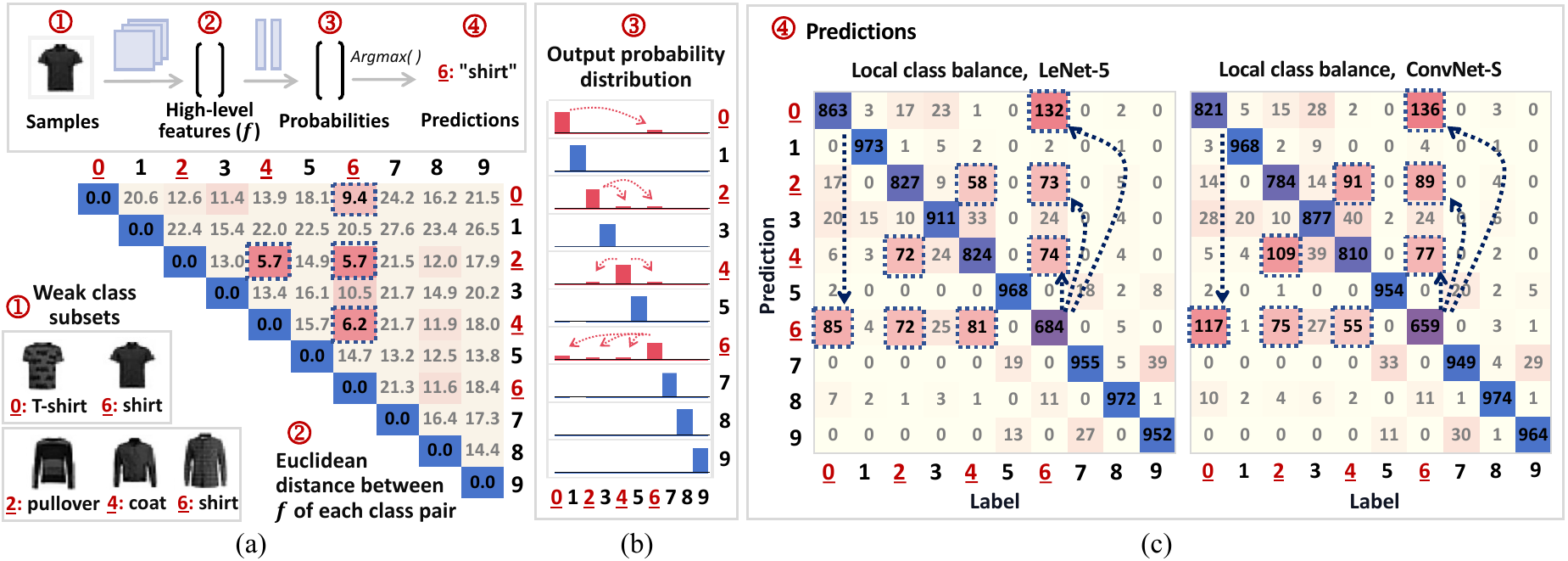} 	
    \caption{
    FashionMNIST: (a) Raw Samples and the Similarities in High-Level Features between each Pair of Classes; (b) Output Probabilities (Averaged over 1000 Samples per Class); (c) Prediction Results. For Simplicity, each Class is Assigned a Serial Number. (0: T-shirt, 1: Trouser, 2: Pullover, 3: Dress, 4: Coat, 5: Sandal, 6: Shirt, 7: Sneaker, 8: Bag, 9: Ankle Boot.)
  }\label{fig: 4}
\end{figure*}

\textbf{Potential reasons for inherent ICD.} Fig.~\ref{fig: 4} (c) maps the predictions to the original classes to which the data belongs (the models are trained once). It shows that the negative samples of a weak class are often assigned certain wrong labels. For the FashionMNIST dataset with a balanced local class distribution, 62\% (=$\frac{85}{137}$) of the misclassified `T-shirt' samples are identified as `shirts' using the LeNet-5 model \cite{MNIST}. Conversely, around 42\% (=$\frac{132}{316}$) of the misclassified `shirt' samples are labeled as `T-shirts'. Another group of classes that can easily be confused with each other include \{`pullover', `coat', and `shirt'\}.
For instance, 23\% of the misclassified `shirt' samples are identified as `pullovers', while 23\% are mistaken for `coats'. That is to say, a total of 88\% of the misclassified `shirt' samples are incorrectly categorized as other weak classes.
This phenomenon becomes particularly pronounced in scenarios with local class imbalance. Approximately 81\% of the overall samples categorized under `pullover' and `coat' are misidentified as `shirt' by the LeNet-5 model. A similar phenomenon is observed with other network models, as shown in Fig.~\ref{fig: 4} (c).

%% Figure 4: 

\begin{figure}[!t]
\setlength{\belowcaptionskip}{-6.5pt}
\centering
    \includegraphics[width=0.49\textwidth]{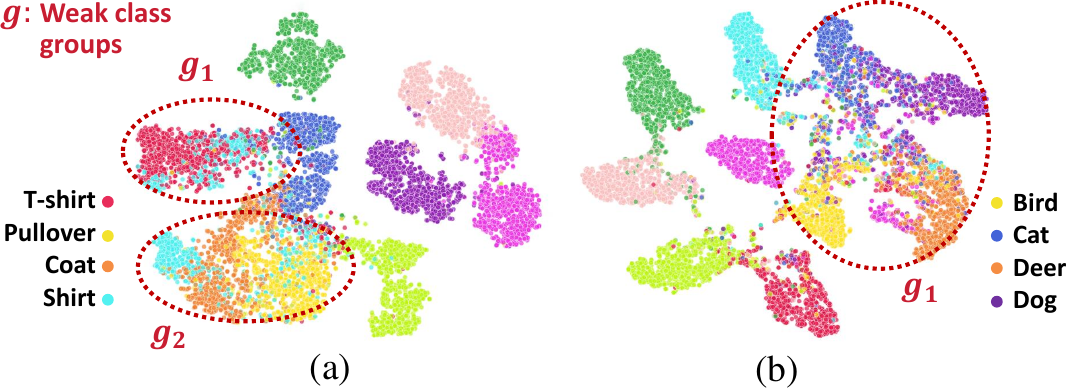} 	
    \caption{t-SNE Visualizations of Feature Vectors on (a) FashionMNIST and (b) CIFAR-10 (under Local Class-Balanced Scenario, 10 clients).}
    \label{fig: 5}
\end{figure}

Fig.~\ref{fig: 4}(a) presents some misclassified samples from the two weak class groups, i.e., \{`T-shirt', `shirt'\} and \{`pullover', `coat', `shirt'\}. It can be seen that these samples exhibit considerable similarities. Fig.~\ref{fig: 4}(a) assesses the similarities in the high-level features ($f$) between each pair of classes. The weak classes within the two aforementioned groups exhibit a lower Euclidean distance compared to the others, indicating a higher degree of similarity. Therefore, it is challenging for the classifier to differentiate among these classes, leading to lower accuracy and reduced confidence in the model's predictions. The probability distributions determined by the softmax function are depicted in Fig.~\ref{fig: 4}(b). The figure displays the average probability distribution across all samples within each class. The t-SNE visualizations of feature vectors are presented in Fig.~\ref{fig: 5}.

%% Figure 5: 

\begin{figure}[!t]
\setlength{\belowcaptionskip}{-4pt}
\centering
    \includegraphics[width=0.48\textwidth]{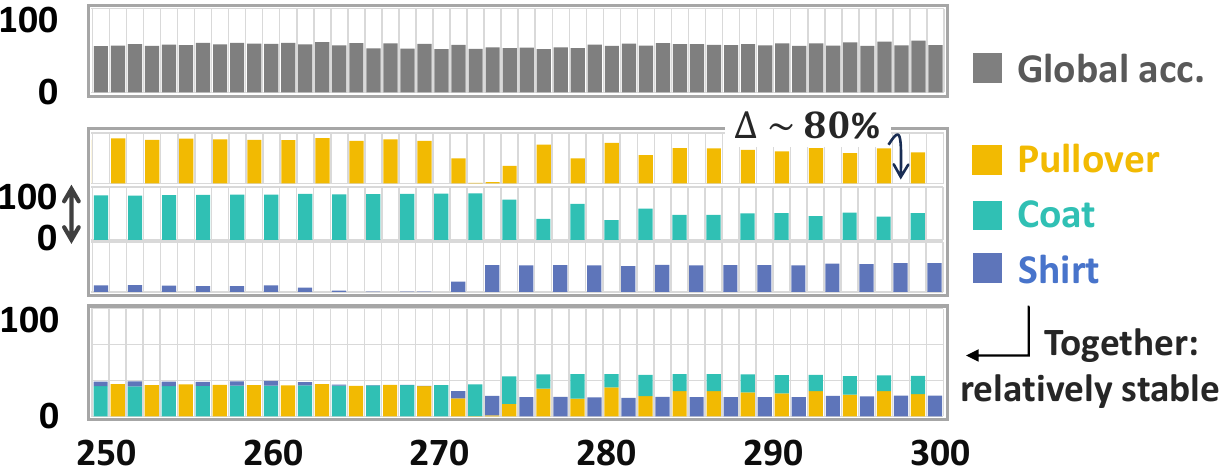} 	
    \caption{The Class-wise Accuracy (\%) During Training (FashionMNIST, under Local Class-Imbalanced Scenario). X Axis: Rounds; Y Axis: Accuracy.}
    \label{fig: 6}
\end{figure}

We further investigate how the classification accuracy for weak classes changes during training. As depicted in Fig.~\ref{fig: 6}, the variance between two consecutive rounds can reach up to 80\% under the local class-imbalanced scenario, even when the global accuracy has become relatively stable.
Furthermore, observations reveal a general trend within the group \{`pullover', `coat', `shirt'\}: an improvement in the accuracy of one class is accompanied by a corresponding decrease in another's.
The collective accuracy across these three classes appears to be relatively stable. The value hovers around $1/3$, implying that the model can generally differentiate weak classes from dominant classes but may fail to recognize the differences among the weak classes themselves.

%% Figure 6: 

While FashionMNIST is used as an example for ease of illustration, similar phenomena are observed in other datasets as well. For instance, in the MNIST dataset, samples from the weak classes `8' and `9' are more likely to be misclassified as \{`3', `5'\} and \{`4', `7'\}, respectively. In the CIFAR-10 dataset, the four weak classes \{`bird', `cat', `deer', `dog'\} are prone to confusion with one another. Interestingly, while these classes may not appear similar in their raw image forms, they exhibit similar high-level features that can make them challenging to distinguish for classifiers.

\section{The Proposed Algorithm}

\subsection{Overview}
\label{sec: 3.1}

\begin{figure}[!t]
  \centering
  \setlength{\belowcaptionskip}{-2pt}
  \setlength{\abovecaptionskip}{10pt}
  \includegraphics[width=0.49\textwidth]{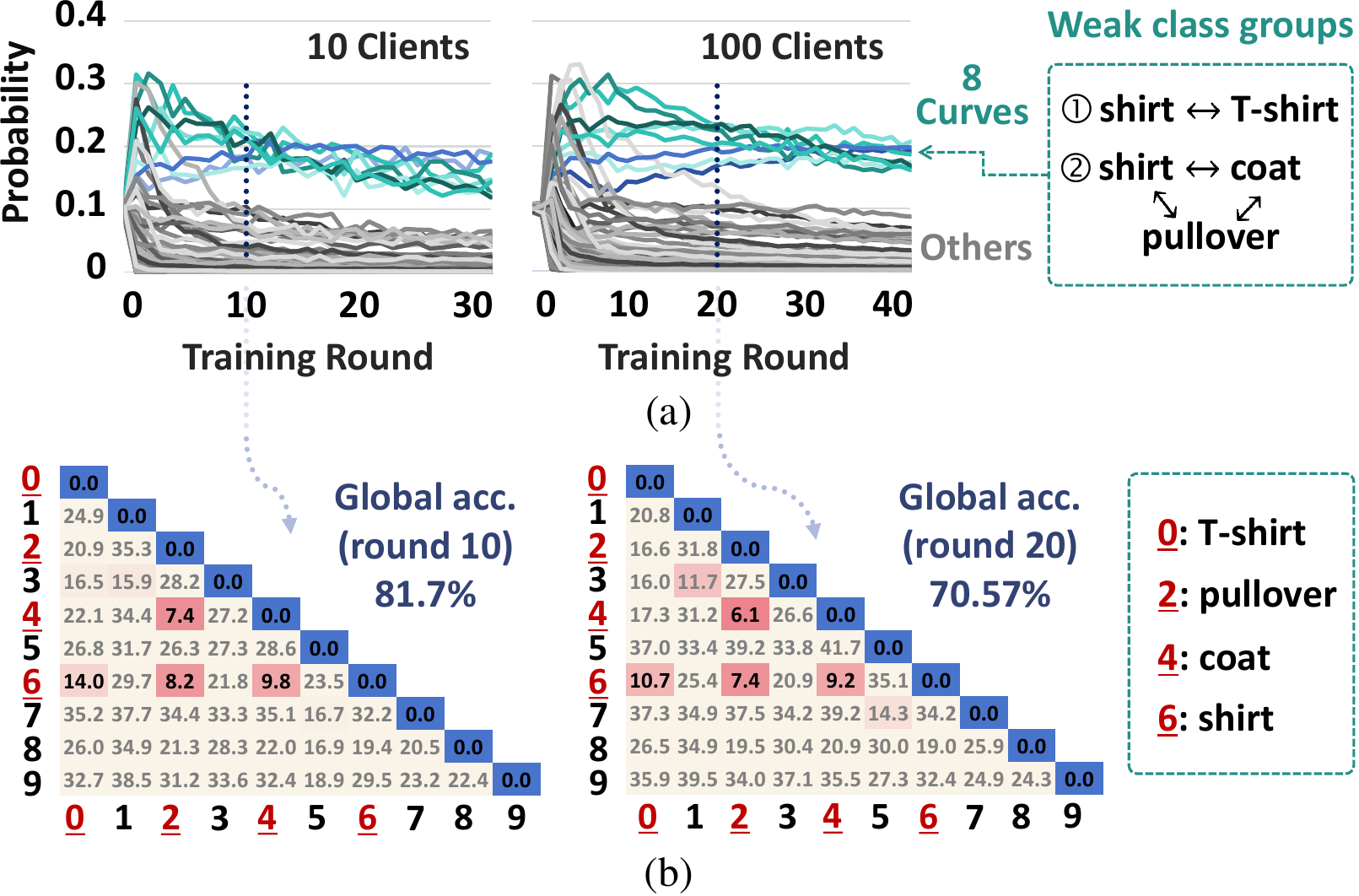} 	
  \caption{(a) The Probability of Samples from One Class Being Misclassified as Another  (FashionMNIST, 10$\times$9 Curves in Total). $~~~$(b) Euclidean Distance between High-Level Features ($f$) of each Class Pair at Round 10$\sim$20.}
  \label{fig: 7}
\end{figure}

In this work, we introduce a federated partial knowledge distillation (PKD) method to mitigate the impact of inherent ICD and enhance the classification accuracy of weak classes. The flow of the proposed method is outlined in Algorithm 1. 
The initial model first undergoes several rounds (e.g., 10 to 20) of federated learning.
After the warmup stage, it enters an expert learning phase (Line 6). 
Weak classes are first identified at the edge by examining the output probability distribution obtained from a feedforward pass on the local training set.
For illustrative purposes, we use the FashionMNIST dataset as an example. 
Fig.~\ref{fig: 7}(a) shows the predicted probabilities for classes other than the true class. Each curve represents the probability of samples from one class being misclassified as another class. As depicted in the figure, after the short warmup phase, the chance of misclassifying one weak class as another within the same group---such as a `T-shirt' being predicted as a `shirt' or vice versa---is significantly higher compared to other cases.
This distinction can help to identify the weak class groups.
Specifically, the eight highlighted curves in Fig.~\ref{fig: 7}(a) correspond to four pairs of weak classes, namely \{`shirt', `T-shirt'\}, \{`shirt', `coat'\}, \{`shirt', `pullover'\}, and \{`coat', `pullover'\}. These four weak class pairs form two groups, as shown in the figure.
Fig.~\ref{fig: 7}(b) further presents the similarities in high-level features ($f$) between each pair of classes at training round 10$\sim$20. It is worth noting that both Fig.~\ref{fig: 7}(a) and (b) reveal pronounced disparities between weak classes and others, even during the early training phase.

%% Algorithm 1: 
\begin{figure}[!t]
  \centering
  \vspace{4pt}
  \includegraphics[width=0.46\textwidth]{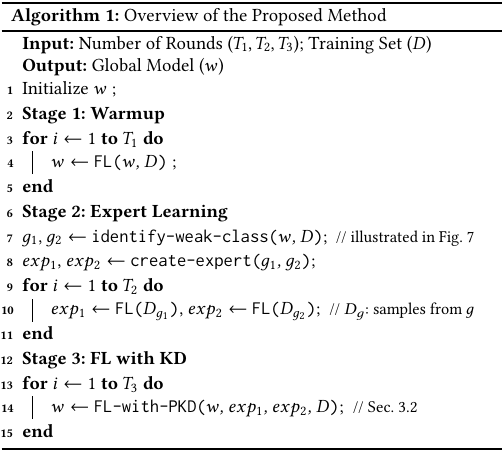} 	
  \label{alg: 1}
\end{figure}
  
%% Figure 7: 

The identified weak class groups are denoted as $g_i$ in Line 7.
For example, in FashionMNIST, $g_1$ consists of \{`T-shirt', `shirt'\}, and $g_2$ comprises of \{`pullover', `coat', and `shirt'\}. In the case of CIFAR-10, there is only one weak class group \{`bird', `cat', `deer' and `dog'\}. Once a group of weak classes is identified ($g_i$), a specialized expert model is trained on a subset of the training data ($D_{g_i}$) to extract and discern the subtle feature differences among them. 
($D_{g_i}$ in Line 10 refers to the samples from class group $g_i$.)
For simplicity, the expert model adopts the same network structure as the global model, with the exception of the last layer. 
The last layer has fewer neurons, as the expert is trained for specific classes.
All experts are trained within the FL framework, ensuring that no data is uploaded to the cloud.
This process takes 25 local training rounds for the FashionMNIST dataset. The overhead of expert training for various benchmarks is analyzed in Section~\ref{sec: 3.3} and Section~\ref{sec: 4.5}.
Typically, one or two experts are adequate to effectively improve the accuracy of the worst-performing classes and alleviate the inherent ICD problem. 

%% Algorithm 2: 
\begin{figure}[!t]
\centering
    \includegraphics[width=0.465\textwidth]{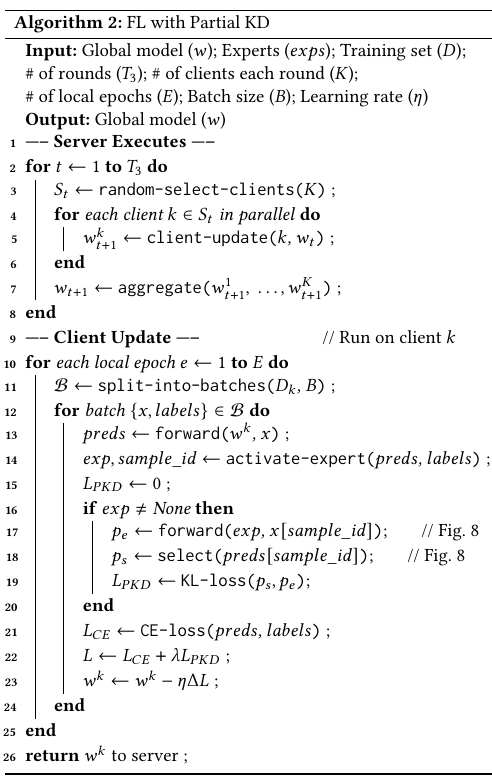} 	
    \label{alg: 2}
\vspace{-9pt}
\end{figure}

In the third stage, partial KD is implemented to transfer the class-specific knowledge from the specialized expert models to the student model, all within the FL framework. This approach is different from the conventional KD methods in several aspects. First, it is triggered only when misclassifications between certain classes (within $g_i$) happens, which means only a subset of misclassified samples is considered in computing the KD loss. Second, while there may be multiple experts, only one expert can be involved in the KD process for any given misclassified sample. Further details regarding the partial KD process will be presented in the following subsection.

\subsection{Partial Knowledge Distillation}
\label{sec: 3.2}

The complete pseudo-code of the PKD process is given in Algorithm~2. The overall framework is similar to the standard baseline, FedAvg \cite{fedavg}, with the exception of the loss calculation described in Line 22. At each round, the global model is broadcast to $K$ randomly selected clients. Each client trains the model locally for $E$ epochs using its own dataset (Line 5). Then, the local models are sent back and aggregated on the server (Line 7) as described in \cite{fedavg}. 

\begin{figure}[!t]
\setlength{\belowcaptionskip}{-4pt}
\centering
    \includegraphics[width=0.485\textwidth]{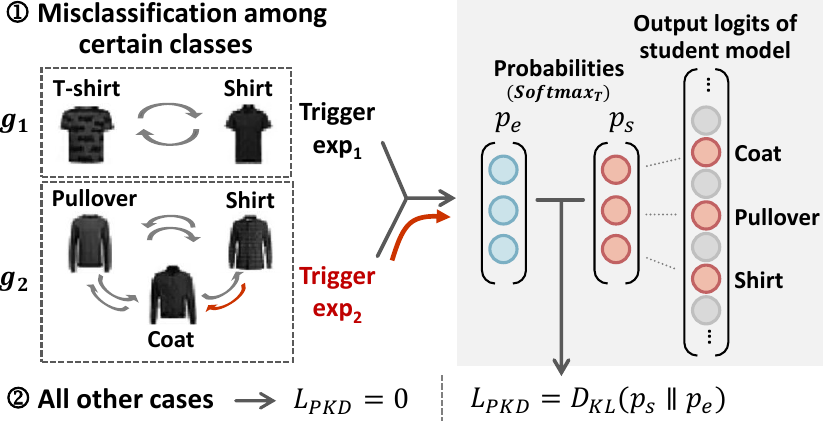} 	
    \caption{An Illustration Example for the Partial KD Process.}
    \label{fig: 8}
\end{figure}

During the local training phase on client $k$, the prediction results ($preds$) are analyzed to detect any instances of misclassification among certain classes in Line 14. 
The indices of these misclassified samples are denoted as $sample\_id$.
An illustration example is presented in Fig.~\ref{fig: 8}. 
Once misclassification within group $g_i$ occurs, the corresponding expert model $exp_i$ is triggered. 
The experts' prediction results on the misclassified samples ($x[{sample\_id}]$) are denoted as $p_e$ (Line 17). 
A PKD loss is then computed using the KL divergence \cite{KLD} between $p_s$ and $p_e$ in Line 19:
\begin{equation} \label{eq: 1}
L_{PKD} = D_{KL}(p_s \parallel p_e) 
\end{equation}
where $p_s$ represents the corresponding predicted probabilities obtained from the student model, as illustrated in Fig.~\ref{fig: 8}. The probabilities $p_e$ and $p_s$ are calculated using the Softmax function with a temperature parameter $T$, formulated as follows:
\begin{equation} \label{eq: 2}
p_i = \frac{e^{z_i/T}}{\sum_j e^{z_j/T}}
\end{equation}
where $z_i$ denotes the output logit. Since the PKD method targets specific classes, $p_s$ and $p_e$ only involve a subset of classes, e.g., 3 classes in Fig.~\ref{fig: 8}. 
In the case where no expert model is activated ($exp$ = None in Line 16), $L_{PKD}$ is simply set to zero. The complete loss function is formulated as Line 22 in Algorithm 2, where $L_{CE}$ denotes the standard cross-entropy loss, and $\lambda$ is a hyperparameter.

%% Figure 8: 

\subsection{Overhead Analysis}
\label{sec: 3.3}

While the proposed method can effectively improve the model's performance for weak classes, extra computational costs may be introduced by the expert training in stage 2 and the partial KD process in stage 3. Assuming the computational effort required for a local training round in the baseline scenario is denoted as $U$. The value can be calculated as follows:
\begin{equation}  \label{eq: 4}
U=E \cdot N \cdot (U_{FP}+U_{BP})
\end{equation}
where $E$ and $N$ denote the number of local epochs and the total number of samples across all $C$ classes, respectively. $U_{FP}$ and $U_{BP}$ correspond to the computational loads for a forward pass and a backward pass, respectively.
The number of computations per round for the proposed method is analyzed in detail as follows.

\vspace{2pt}
Stage 1: Warmup
\vspace{-2pt}
\begin{equation}  \label{eq: 5}
U_{T_1}=U
\end{equation}

\vspace{-2pt}
Stage 2: Expert learning
\vspace{-2pt}
\begin{equation} \label{eq: 6}
U_{T_2}=E_{exp} \cdot \sum_{i} n_{g_i} \cdot (U_{FP}+U_{BP})= \frac{E_{exp}}{E} \cdot \frac{\sum_{i} n_{g_i}}{N} \cdot U
\end{equation} 

\vspace{-1pt}
Stage 3: FL with PKD
\vspace{-4pt}
\begin{equation} \label{eq: 7}
U_{T_3}= U + n_{KD}^t \cdot U_{FP} \approx (1 + \frac{1}{3} \cdot \frac{n_{KD}^t}{E \cdot N}) \cdot U
\end{equation}
where $E_{exp}$ denotes the number of local epochs for expert training (usually equal to $E$), and $n_{g_i}$ represents the number of samples belonging to the class group $g_i$. The value of $\sum_{i} n_{g_i}$ ranges from 0.09$N$ to 0.6$N$ for the datasets used in this work. 
$~n_{KD}^t$ in Eq.~\ref{eq: 7} denotes the number of misclassified samples that activate the experts in round~$t$.  $~~\frac{n_{KD}^t}{E\cdot N}$ turns out to be a small value ranging from 0.003 to 0.16, indicating that the expert models are only partially involved in the FL process. 
Compared to the conventional KD methods \cite{FedCodl,KDrelevant2}, the overall computational overhead of the proposed PKD method is significantly reduced. This reduction is attributed to the limited number of samples that participate in expert learning and the PKD process. We will further demonstrate that with the same computation cost, the proposed method can achieve higher accuracy than the baseline method.

\vspace{6pt}
\section{Experimental Results}
\subsection{Experimental Setups}
\label{sec: 4.1}

\textbf{Datasets.} We evaluate the proposed method on several widely-used datasets in FL, including the simple datasets MNIST and FashionMNIST, as well as the more complex datasets CIFAR-10 and CIFAR-100 \cite{Cifar10}. We also adopt the more realistic and naturally partitioned dataset FEMNIST \cite{FEMNIST} for evaluation.  

\textbf{Data partitioning strategies.} To make a comprehensive analysis, we employ four data partitioning strategies. 
1) \textit{Local class-balanced partition}: Each client's dataset maintains strictly equal sample sizes across all classes, ensuring uniform local class distribution. 
2) \textit{Pathological partition} \cite{fedavg}: Each client contains data samples from only a small subset of classes (e.g., 1-2 classes), leading to severe distribution skew. 
Unless otherwise stated, we adopt a single-class-per-client assignment by default to ensure a clear and straightforward comparison of class-wise accuracy.
3) \textit{The Dirichlet partition} \cite{DirichletFL}: 
Sample allocation follows a Dirichlet distribution $Dir(\alpha)$ \cite{DirichletDistribution} to generate a range of non-i.i.d. data scenarios, enabling comprehensive analysis of the inherent ICD phenomenon and evaluation of the PKD method under varying heterogeneity levels. 
4)~\textit{FEMNIST's Native Partition}: The dataset's native writer-based partition is preserved to simulate practical FL scenarios.
Global class balance across all clients is preserved in all experiments except for FEMNIST, which maintains its natural data distribution.

\textbf{Models.} Similar to \cite{fedavg,FedMA,WeightAggregation}, LeNet-5 is adopted for the MNIST, FashionMNIST, and FEMNIST datasets by default. We also use the same lightweight network ConvNet-S as in \cite{fedavg} for comparison. (ConvNet-L is simply an extended version of ConvNet-S with an additional 5$\times$5 convolutional layer.)
For CIFAR-10 and CIFAR-100, we use ResNet-56 \cite{ResNet} by default and also evaluate with the shallower network VGG-9 \cite{vgg,FedMA}. 
As mentioned in Section~\ref{sec: 3.1}, the expert model adopts the same network structure as the global model, with the exception of the last layer. The number of output neurons is equal to the size of the weak class subset.
The number of experts aligns with the number of weak class subsets detailed in Section~\ref{sec: 2.2}: two for MNIST and FashionMNIST, and one for CIFAR-10. Since CIFAR-100 comprises 100 classes, multiple weak class groups exist. We select the $G$ worst-performing groups based on the lowest average accuracy.

% \documentclass{article}
% \usepackage{multirow}
% \usepackage{booktabs}
% \usepackage{graphicx} % for \resizebox

% \begin{document}

\begin{table}[t]
  \centering
  \setlength{\tabcolsep}{2.5pt}  
  \caption{Performance of PKD under the Local Class-Balanced Scenario}\label{tab_tex:1}
  \makebox[\linewidth][c]{
    \resizebox{1.0\linewidth}{!}{
      \begin{tabular}{cc ccc ccc ccc}
      \toprule

      \multirow{2}{*}{\textbf{\small  \#Client}} & \textbf{\small Dataset}
      & \multicolumn{3}{c}{\textbf{MNIST}} 
      & \multicolumn{3}{c}{\textbf{FashionMNIST}} 
      & \multicolumn{3}{c}{\textbf{CIFAR-10}} \\

      \cmidrule(lr){2-2} \cmidrule(lr){3-5} \cmidrule(lr){6-8} \cmidrule(lr){9-11}
        & \textbf{\small Method}
        & \textbf{FedAvg} & \textbf{PKD} & \textbf{$\Delta$} 
        & \textbf{FedAvg} & \textbf{PKD} & \textbf{$\Delta$}  
        & \textbf{FedAvg} & \textbf{PKD} & \textbf{$\Delta$} \\

      \midrule
      \multirow{4}{*}{\textbf{10}} 
      & \textbf{Max.} & 99.73 & 99.73 & \textbf{0.00} & 97.30 & 97.60 &  \textbf{0.30} & 93.10 & 93.00 &  \textbf{-0.10} \\
      & \textbf{Ave.} & 99.28 & 99.32 & \textbf{0.04} & 89.29 & 89.50 &  \textbf{0.21} & 83.78 & 84.93 &  \textbf{1.15} \\
      & \textbf{Min.} & 98.31 & 98.71 & \textbf{0.40} & 68.40 & 76.60 &  \textbf{8.20} & 59.60 & 71.90 &  \textbf{12.30} \\
       \cmidrule(lr){2-11}
      & \textbf{Worst} &``9''&``9''& &``Shirt''&``Shirt''& &``Cat''&``Cat''& \\

      \midrule
      \multirow{4}{*}{\textbf{100}} 
      & \textbf{Max.} & 99.28 & 99.29 & \textbf{0.01} & 96.70 & 96.60 & \textbf{-0.10} & 80.90 & 82.70 & \textbf{1.80} \\
      & \textbf{Ave.} & 98.11 & 98.34 & \textbf{0.23} & 84.66 & 84.72 &  \textbf{0.06} & 68.52 & 70.11 & \textbf{1.59} \\
      & \textbf{Min.} & 96.72 & 96.92 & \textbf{0.20} & 51.60 & 62.30 & \textbf{10.70} & 46.70 & 53.00 & \textbf{6.30} \\
      \cmidrule(lr){2-11}
      & \textbf{Worst} &``9''&``9''& &``Shirt''&``Shirt''& &``Cat''&``Cat''& \\
      \bottomrule
      \end{tabular}
    }
  }
\end{table}

% \end{document}

\textbf{Compared Methods.} In this section, we compare the performance of the PKD method with the standard baseline, FedAvg, as well as several prior techniques \cite{Focal, BalanceFL, cutout,FedProx,FedCodl} under various data partitioning setups. 
These advanced techniques include focal loss \cite{Focal} and data augmentation \cite{BalanceFL,cutout}, both designed to mitigate the ICD resulting from class imbalance.
We also compare with two other methods, FedProx \cite{FedProx} and FedCodl \cite{FedCodl}, under the Dirichlet partitions. These methods are designed to alleviate the data heterogeneity problem in FL by either constraining local updates or incorporating global knowledge into the local training process through KD.  
Our analysis reveals that while current approaches effectively alleviate general data heterogeneity, they remain insufficient in addressing the inherent ICD problem.
   
\textbf{Hyperparameters.} Stochastic Gradient Descent (SGD) is employed for optimization, with a learning rate of 0.01.
Same as FedAvg~\cite{fedavg}, the number of local epochs and the batch size are set to 5 and 50, respectively.  
The networks are trained for 100 rounds on MNIST and up to 300 rounds on other datasets.
The total number of clients is set to 10 by default. We also scale this default value to 100 across various benchmarks and vary the fraction of clients selected at each round to evaluate PKD under different setups. Three fractions are used on the FEMNIST dataset, including 0.1, 0.5, and 1.0.
During the PKD process, the hyperparameter $T$ is set to 5. The value of $\lambda$ is set to 1.0 for MNIST, FashionMNIST, and FEMNIST, and to 0.5 for CIFAR-10 and CIFAR-100.  

\textbf{Implementation.} All experiments are implemented in Python using the PyTorch framework. Training is conducted on a server running Ubuntu, equipped with 4 NVIDIA RTX 3090 GPUs, 2 Intel Xeon Platinum 8269CY CPUs, and 128 GB of RAM.

\subsection{Performance of the Proposed Method}
\label{sec: 4.2}

% \documentclass{article}
% \usepackage{multirow}
% \usepackage{booktabs}
% \usepackage{graphicx} % for \resizebox

% \begin{document}
% \small
\begin{table}[t]
    \centering
    \setlength{\tabcolsep}{9pt}  
    \caption{Results for Other Neural Networks under the Local Class-Balanced Scenario}\label{tab_tex:2}  
    \makebox[\linewidth][c]{
      \resizebox{1.0\linewidth}{!}{
        \begin{tabular}{c  ccc ccc}
          \toprule

          \textbf{Dataset} &\multicolumn{3}{c}{\textbf{FashionMNIST}} & \multicolumn{3}{c}{\textbf{CIFAR-10}} \\
          \cmidrule{1-1}  \cmidrule(lr){2-4}  \cmidrule(lr){5-7} 
          \textbf{Network} &\multicolumn{3}{c}{\textbf{ConvNet-S}} & \multicolumn{3}{c}{\textbf{VGG-9}} \\
          \cmidrule{1-1} \cmidrule(lr){2-4}  \cmidrule(lr){5-7} 
          \textbf{Method} & \textbf{FedAvg} & \textbf{PKD} & \textbf{$\Delta$} & \textbf{FedAvg} & \textbf{PKD} & \textbf{$\Delta$}\\ 
          
          \midrule
          \textbf{Max.} & 98.20 & 98.20 & \textbf{0.00} & 92.50 & 92.60 & \textbf{0.10} \\
          \textbf{Ave.} & 90.43 & 90.92 & \textbf{0.49} & 83.04 & 83.48 & \textbf{0.44} \\
          \textbf{Min.} & 70.50 & 75.30 & \textbf{4.80} & 65.10 & 72.30 & \textbf{7.20} \\
          
          \cmidrule(lr){1-7}
          \textbf{Worst} &``Shirt''&``Shirt'' & &``Cat''&``Cat''&\\ %& &``Cat''&``Cat''& \\
          \bottomrule  
        \end{tabular}

      }
    }
\end{table}

% \normalsize
% \end{document}

Table~\ref{tab_tex:1} presents the performance of our proposed method under \textit{local class-balanced scenario} as described in Section~\ref{sec: 4.1}.
In the baseline scenario with 10 clients, the inter-class accuracy discrepancy is 1.42\%, 28.9\% and 33.5\% for MNIST, FashionMNIST and CIFAR-10, respectively. 
The proposed PKD method can improve the accuracy of the worst-performing classes by 0.4\%, 8.2\% and 12.3\%, respectively, without comprising the global average accuracy. 
The overall model performance degrades as the number of clients increases to 100. In this case, PKD can improve the worst class-wise accuracy by 0.2\%, 10.7\%, and 6.3\% on MNIST, FashionMNIST, and CIFAR-10, respectively.
It is worth noting that improving the accuracy of weak classes usually does not compromise the model performance on other classes. The accuracy of dominant classes remains largely unchanged or is even slightly improved. As a result, the global average accuracy also increases. 
As mentioned in Section~\ref{sec: 4.1}, we employ LeNet-5 as the baseline architecture for evaluations on MNIST and FashionMNIST, while utilizing ResNet-56 for CIFAR-10. 
For a comprehensive comparison, Table~\ref{tab_tex:2} further presents the performance of PKD for additional network architectures.

\begin{table}[t]
    \centering
    \setlength{\tabcolsep}{3pt}  
    \caption{Performance of PKD under the Pathological Data Partition Scenario}\label{tab_tex:3}  
    \makebox[\linewidth][c]{
      \resizebox{1.0\linewidth}{!}{
        
        \begin{tabular}{c ccc ccc ccc}
            \toprule
            \textbf{Dataset} & \multicolumn{3}{c}{\textbf{MNIST}} & \multicolumn{3}{c}{\textbf{FashionMNIST}} & \multicolumn{3}{c}{\textbf{CIFAR-10}} \\
            \cmidrule(lr){1-1} \cmidrule(lr){2-4} \cmidrule(lr){5-7} \cmidrule(lr){8-10}
            \textbf{\small \#Class/Client} & \multicolumn{3}{c}{\textbf{1}} & \multicolumn{3}{c}{\textbf{1}} & \multicolumn{3}{c}{\textbf{2}} \\
            \cmidrule(lr){1-1} \cmidrule(lr){2-4} \cmidrule(lr){5-7} \cmidrule(lr){8-10}
            \textbf{Method} & \textbf{FedAvg} & \textbf{PKD} & \textbf{$\Delta$} & \textbf{FedAvg} & \textbf{PKD} & \textbf{$\Delta$} & \textbf{FedAvg} & \textbf{PKD} & \textbf{$\Delta$} \\
            \midrule
            \textbf{Max.} & 96.02 & 99.03 & \textbf{3.01} & 91.20 & 91.40 & \textbf{0.20} & 72.10 & 80.30 & \textbf{8.20} \\
            \textbf{Ave.} & 75.85 & 84.45 & \textbf{8.60} & 47.39 & 71.03 & \textbf{23.64} & 38.64 & 41.42 & \textbf{2.78} \\
            \textbf{Min.} & 7.90 & 68.38 & \textbf{60.48} & 0.00 & 51.80 & \textbf{51.80} & 2.90 & 16.10 & \textbf{13.20} \\
            \cmidrule(lr){1-10}
            \textbf{Worst}  &``8''&``9''& &``Shirt''&``Shirt''& &``Bird''&``Cat''& \\

            \bottomrule
        \end{tabular}

      }
    }
\end{table}

\begin{figure}[t]
  \setlength{\belowcaptionskip}{-2pt}
  \centering
      \includegraphics[width=0.48\textwidth]{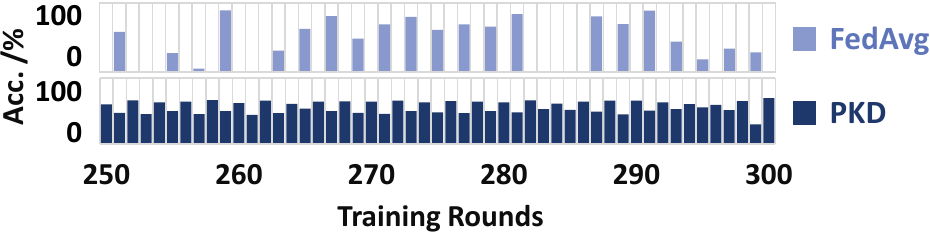} 	
      \caption{Accuracy of the Worst-performing Class `Shirt' during Training. (FashionMNIST, \#Class/Client=1)}
      \label{fig: 9}
\end{figure}

Table~~\ref{tab_tex:3} shows the performance of PKD under the \textit{pathological partition scenario}.
The number of classes per client (\#Class/Client) is set to 1 for MNIST and FashionMNIST, and to 2 for CIFAR-10. 
Similar to the results reported in \cite{extremeNonIID}, the global accuracy becomes low under such an extremely unbalanced distribution.
Compared to local class-balanced scenarios, performance improvements for the worst-performing classes are more pronounced, as shown in Table~~\ref{tab_tex:3}.
The minimum class-wise accuracy increases by 60.5\% on MNIST, 51.8\% on FashionMNIST, and 13.2\% on CIFAR-10, respectively.
As a result, the global average accuracy is also improved.
As shown in Fig.~\ref{fig: 9}, the accuracy of weak classes becomes significantly more stable compared to the baseline case.

Table~~\ref{tab_tex:4} presents the experimental results under the \textit{Dirichlet partition scenario}. 
The concentration parameter $\alpha$ in $Dir(\alpha)$ is set to 0.5. The overall model performance is better than in the pathological partition scenario. However, the baseline methods still exhibit 3.01\% inherent ICD on MNIST, 43.8\% on FashionMNIST, and 37.6\% on CIFAR-10.
Our PKD approach improves the minimum class-wise accuracy by 0.39\%, 10.10\%, and 6.10\%, respectively, without affecting the average accuracy.

\begin{table}[t]
    \centering
    \setlength{\tabcolsep}{3pt} 
    \caption{Performance of PKD under the Dirichlet Partition Scenario ($Dir~(0.5)$)}\label{tab_tex:4} 
    \makebox[\linewidth][c]{
      \resizebox{1.0\linewidth}{!}{
        
        \begin{tabular}{c ccc ccc ccc}
            \toprule
            \textbf{Dataset} & \multicolumn{3}{c}{\textbf{MNIST}} & \multicolumn{3}{c}{\textbf{FashionMNIST}} & \multicolumn{3}{c}{\textbf{CIFAR-10}} \\
            \cmidrule(lr){1-1} \cmidrule(lr){2-4} \cmidrule(lr){5-7} \cmidrule(lr){8-10}
            \textbf{Method} & \textbf{FedAvg} & \textbf{PKD} & \textbf{$\Delta$} & \textbf{FedAvg} & \textbf{PKD} & \textbf{$\Delta$} & \textbf{FedAvg} & \textbf{PKD} & \textbf{$\Delta$} \\
            \midrule
            \textbf{Max.} & 99.64 & 99.64 & \textbf{0.00} & 96.90 & 96.90 & \textbf{0.00} & 92.90 & 93.80 & \textbf{0.90} \\
            \textbf{Ave.} & 98.72 & 98.84 & \textbf{0.12} & 86.05 & 86.20 & \textbf{0.15} & 76.38 & 77.92 & \textbf{1.54} \\
            \textbf{Min.} & 96.63 & 97.02 & \textbf{0.39} & 53.10 & 63.20 & \textbf{10.10} & 55.30 & 61.40 & \textbf{6.10} \\        
            \cmidrule(lr){1-10}
            \textbf{Worst} & ``9'' & ``9'' & & ``Shirt'' & ``Shirt'' & & ``Dog'' & ``Cat'' & \\

            \bottomrule
        \end{tabular}

      }
    }
\end{table}

%% Figure 9: 

Table~~\ref{tab_tex:5} presents the results on the FEMNIST dataset based on its native partition. The dataset comprises of over 3000 users and 62 classes, including digits, uppercase, and lowercase letters.
To make a direct comparison with the MNIST dataset, we retain only the digit samples and their corresponding users.
In realistic FL scenarios, not all users participate in every communication round. To simulate this, we set the fraction of clients selected at each round to 0.1, 0.5, and 1.0, respectively. This participation fraction applies to both the expert learning and the global model learning processes.
PKD shows consistent improvements across different fractions, increasing the minimum class-wise accuracy by 1.4\%, 1.5\%, and 1.7\%, respectively.

\subsection{Comparison with Existing Techniques}
\label{sec: 4.3}

Table~\ref{tab_tex:6} compares the proposed method with several existing techniques that tackle the ICD problem induced by class imbalance.
An effective approach to address the class imbalance issue is to offset the negative impact brought by the uneven class distribution \cite{CBloss, Focal}. It can be done by either re-sampling the minority classes or assigning larger weights to their loss values while calculating the empirical risk. Here we adopt a similar idea as \cite{Focal} and test its performance for alleviating the inherent ICD problem. Instead of up-weighting the minority classes, we dynamically enhance the weights of the inherent weak classes when calculating the loss value. 
The result indicates that directly applying this method may not alleviate the inherent ICD problem. 
It is observed that the minimum class-wise accuracy drops due to increased disparity among weak classes within a group.
Take the weak class group \{`shirt', `coat', `pullover'\} in Fig.~\ref{fig: 6} as an example. As mentioned in Section~\ref{sec: 2.2}, an improvement in the accuracy of one weak class is generally accompanied by a corresponding decrease in the accuracy of other weak classes. Suppose at a given epoch, `pullover' is the worst-performing class and assigned a large weight while calculating the total loss. 
In subsequent iterations, the model is updated to focus more on improving performance for `pullover'. However, this improvement may come at the cost of degraded performance on the other two weak classes, thereby resulting in no significant improvement in overall accuracy. 
In \cite{cutout} and \cite{BalanceFL}, data augmentation is applied at the sample level and feature level, respectively, to implicitly enhance data diversity. These two approaches are commonly used to improve the model's performance on minority classes. The results suggest that feature-level data augmentation can modestly mitigate the inherent ICD issue, leading to a 2.30\% improvement in the accuracy of the worst-performing class.

\begin{table}[t]
    \centering
    \setlength{\tabcolsep}{3pt} 
    \caption{Experimental Results on the FEMNIST Dataset}\label{tab_tex:5} 
    \makebox[\linewidth][c]{
      \resizebox{1.0\linewidth}{!}{
        
        \begin{tabular}{c ccc ccc ccc}
            \toprule
            \textbf{Dataset} & \multicolumn{9}{c}{\textbf{FEMNIST}} \\
            \midrule
            \textbf{Fraction} & \multicolumn{3}{c}{\textbf{0.1}} & \multicolumn{3}{c}{\textbf{0.5}} & \multicolumn{3}{c}{\textbf{1.0}} \\
            \cmidrule(lr){1-1} \cmidrule(lr){2-4} \cmidrule(lr){5-7} \cmidrule(lr){8-10}
            \textbf{Method} & \textbf{FedAvg} & \textbf{PKD} & \textbf{$\Delta$} & \textbf{FedAvg} & \textbf{PKD} & \textbf{$\Delta$} & \textbf{FedAvg} & \textbf{PKD} & \textbf{$\Delta$} \\
            \midrule
            \textbf{Max.} & 98.10 & 98.10 & \textbf{0.00} & 98.10 & 98.20 & \textbf{0.10} & 98.10 & 98.20 & \textbf{0.10} \\
            \textbf{Ave.} & 92.30 & 92.56 & \textbf{0.26} & 92.78 & 93.00 & \textbf{0.22} & 92.96 & 93.27 & \textbf{0.31} \\
            \textbf{Min.} & 87.50 & 89.20 & \textbf{1.70} & 88.20 & 89.70 & \textbf{1.50} & 88.60 & 90.00 & \textbf{1.40} \\            
            \bottomrule
        \end{tabular}

      }
    }
\end{table}

% \documentclass{article}
% \usepackage{multirow}
% \usepackage{booktabs}
% \usepackage{graphicx} % for \resizebox

% \begin{document}

\begin{table}[t]
    \centering
    \setlength{\tabcolsep}{2pt} 
    \caption{Comparison with Existing ICD Solutions for the Class Imbalance Problem.
}\label{tab_tex:6} 
    \makebox[\linewidth][c]{
      \resizebox{1.0\linewidth}{!}{
        \begin{tabular}{cc cc cc cc cc}
          \toprule
          \textbf{Setup} &\multicolumn{9}{c}{\textbf{FashionMNIST, Local Class Balance, \#client=10}} \\
          \midrule
          \multirow{3}{*}{\textbf{Method}} & \multirow{2}{*}{\textbf{Fedavg}} & \multicolumn{2}{c}{\textbf{Re-weighting}} &\multicolumn{4}{c}{\textbf{Data Augmentation}} & \multicolumn{2}{c}{\multirow{2}{*}{\textbf{Ours}}} \\
          \cmidrule(lr){3-4} \cmidrule(lr){5-8} 
          &  & \multicolumn{2}{c}{\textbf{Focal loss}} &\multicolumn{2}{c}{\textbf{Sample-level}} &\multicolumn{2}{c}{\textbf{Feature-level}} & & \\
          \cmidrule(lr){2-2} \cmidrule(lr){3-4} \cmidrule(lr){5-6} \cmidrule(lr){7-8} \cmidrule(lr){9-10}
          & \cite{fedavg} & \cite{Focal} & $\Delta$ & \cite{cutout} & $\Delta$ & \cite{BalanceFL} & $\Delta$ & PKD & $\Delta$ \\
          \midrule
          \textbf{Max.} & 97.30 & 97.80 & \textbf{0.50} & 97.40 & \textbf{0.10} & 97.40 & \textbf{0.10} & 97.60 & \textbf{0.30} \\
          \textbf{Ave.} & 89.29 & 88.64 & \textbf{-0.65} & 85.04 & \textbf{-4.25} & 89.52 & \textbf{0.23} & 89.50 & \textbf{0.21} \\
          \textbf{Min.} & 68.40 & 65.40 & \textbf{-3.00} & 53.80 & \textbf{-14.60} & 70.70 & \textbf{2.30} & 76.60 & \textbf{8.20} \\
          \bottomrule
        \end{tabular}

      }
    }
\end{table}

% \end{document}

Table~\ref{tab_tex:7} compares the proposed method with FedProx \cite{FedProx} and FedCodl \cite{FedCodl} under the Dirichlet partition scenario. These methods are designed to tackle data heterogeneity in FL. 
Specifically, FedProx introduces a proximal term to the objective to constrain local updates and uses a hyperparameter $\mu$  to control its strength.  
FedCodl, on the other hand, adopts a KD framework, encouraging local models to align with the global model by minimizing the KL divergence between their output logits. 
Experimental results show that while existing methods can mitigate general data heterogeneity and improve overall model performance, they exhibit limited effectiveness in enhancing accuracy for weak classes compared to the proposed approach.

% \documentclass{article}
% \usepackage{multirow}
% \usepackage{booktabs}
% \usepackage{graphicx} % for \resizebox

% \begin{document}

\begin{table}[t]
    \centering
    \setlength{\tabcolsep}{3pt} 
    \caption{Comparison with Existing Data-Heterogeneity Solutions}\label{tab_tex:7}  
    \makebox[\linewidth][c]{
      \resizebox{1.0\linewidth}{!}{
        \begin{tabular}{c c cccc ccc cc}
          \toprule
          \textbf{Setup} &\multicolumn{9}{c}{\textbf{FashionMNIST, Dir(0.5), \#client=20}} \\
          \midrule
          \multirow{2}{*}{\textbf{Method}} & \textbf{Fedavg} & \multicolumn{4}{c}{\textbf{FedProx\cite{FedProx}}} &\multicolumn{2}{c}{\textbf{FedCodl}} & \multicolumn{2}{c}{\textbf{Ours}} \\
          \cmidrule(lr){2-2} \cmidrule(lr){3-6} \cmidrule(lr){7-8}  \cmidrule(lr){9-10} 
          & \cite{fedavg} & $\mu = 0.001$ & $\Delta$ & $\mu = 0.002$ & $\Delta$ & \cite{FedCodl} & $\Delta$ &  PKD & $\Delta$ \\
          \midrule
          \textbf{Max.} & 96.90 & 97.00 & \textbf{0.10} & 97.30 & \textbf{0.40} & 97.50 & \textbf{0.60} & 96.90 & \textbf{0.00} \\
          \textbf{Ave.} & 86.05 & 86.17 & \textbf{0.12} & 86.06 & \textbf{0.01} & 86.59 & \textbf{0.54} & 86.20 & \textbf{0.15} \\
          \textbf{Min.} & 53.10 & 56.90 & \textbf{3.80} & 55.60 & \textbf{2.50} & 53.80 & \textbf{0.70} & 63.20 & \textbf{10.10} \\
          \bottomrule
        \end{tabular}

      }
    }
\end{table}

% \end{document}

\subsection{Performance on More Complex Datasets}
\label{sec: 4.4}

\begin{figure}[!t]
\setlength{\belowcaptionskip}{-4pt}
\centering
    \includegraphics[width=0.33\textwidth]{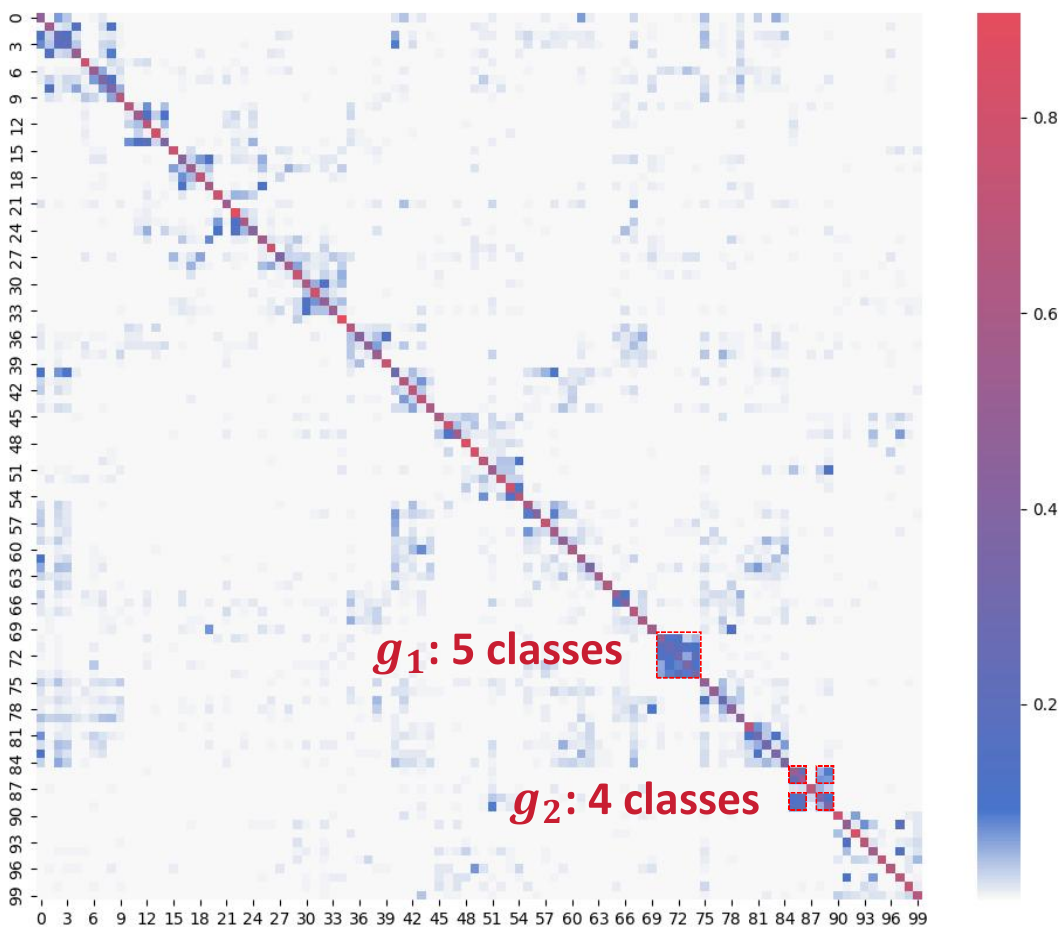} 	
    \caption{Prediction Results of ResNet-56 on CIFAR-100.}
    \label{fig: 10}
\end{figure}

To assess the effectiveness of this approach on more complex datasets with higher class diversity, we further evaluate its performance on CIFAR-100, which comprises 100 classes. We maintain a local class-balanced partition across 10 clients and compare PKD to the baseline. 
Similar to Fig.~\ref{fig: 4}(c) for FashionMNIST, Fig.~\ref{fig: 10} maps the predictions to their respective original classes. Given that CIFAR-100 has a large number of classes, the weak class groups are expected to increase in size.
For the PKD implementation, we select $G$ groups with the poorest average accuracy. Due to constraints on the number of experts and corresponding overhead, $G$ is set to a small value, specifically 1 or 2, in line with other datasets. The number of weak classes in the two groups is 5 and 4, respectively. 
The results are shown in Table~\ref{tab_tex:8}. 
Despite utilizing only 1-2 expert models, the PKD method achieves a 3-5\% improvement in minimum class-wise accuracy.

% \documentclass{article}
% \usepackage{multirow}
% \usepackage{booktabs}
% \usepackage{graphicx} % for \resizebox

% \begin{document}
% \small
\begin{table}[t]
    \centering
    \setlength{\tabcolsep}{10pt} 
    \caption{Experimental Results on the CIFAR-100 Dataset.}\label{tab_tex:8}  
    \makebox[\linewidth][c]{
      \resizebox{1.0\linewidth}{!}{
        \begin{tabular}{cc cc cc}
          \toprule

          \textbf{Setup} &\multicolumn{5}{c}{\textbf{CIFAR-100, Local Class Balance, \#client=10}} \\
          \midrule
          \multirow{2}{*}{\textbf{Method}}& \multirow{2}{*}{\textbf{FedAvg}}& \multicolumn{4}{c}{\textbf{PKD} }\\
          \cmidrule(lr){3-6}
        
          % & &   & \textbf{PKD} & \multirow{2}{*}{\textbf{$\Delta$}} \\
          %  & & & & & \\
          &&\textbf{$G=1$}& \textbf{$\Delta$}& \textbf{$G=2$}&  \textbf{$\Delta$}\\
          \midrule
          \textbf{Max.} & 91.00 & 92.00 & \textbf{1.00} & 92.00 & \textbf{1.00} \\
          \textbf{Ave.} & 62.82 & 64.02 & \textbf{1.20} & 62.91 & \textbf{0.09} \\
          \textbf{Min.} & 27.00 & 30.00 & \textbf{3.00} & 32.00 & \textbf{5.00} \\
          \bottomrule
        \end{tabular}

      }
    }
\end{table}

% \normalsize
% \end{document}

%% Figure 10: 

\subsection{Overhead Analysis}
\label{sec: 4.5}

% \documentclass{article}
% \usepackage{multirow}
% \usepackage{booktabs}
% \usepackage{graphicx} % for \resizebox

% \begin{document}

\begin{table}[t]
    \centering
    \caption{Computation Cost in the Local Class-Balanced Scenario (Unit: $U$, which is 374.87 GFLOPs for the Benchmark)}
    \label{tab_tex:9}
    \makebox[\linewidth][c]{
      \resizebox{1.0\linewidth}{!}{
        \begin{tabular}{c ccccc cc}   % 7c
          \toprule
          \textbf{Setup} & \multicolumn{7}{c}{\textbf{FashionMNIST, Local Class Balance, \#client=10}} \\
          \midrule
          \multirow{2}{*}{\textbf{Method}} & \multicolumn{5}{c}{\textbf{Target Min. Acc.}}& \multicolumn{2}{c}{\textbf{Final Acc.}} \\
          \cmidrule(lr){2-6} \cmidrule(lr){7-8}
           & 40\% & 50\% & 60\% & 65\% & 70\% & \textbf{Min.} & \textbf{Ave.} \\
          \midrule
          \textbf{FedAvg~\cite{fedavg}} & 13 & 19 & 30 & 57 & - & \textbf{68.4\%} & \textbf{89.3\%} \\
          \textbf{Focal Loss~\cite{Focal}} & 14 & 23 & 56 & 72 & - & \textbf{65.4\%} & \textbf{88.6\%} \\
          \textbf{Data Aug.~\cite{cutout}} & 29 & 67 & - & - & - & \textbf{53.8\%} &  \textbf{85.0\%}\\
          \textbf{Data Aug.~\cite{BalanceFL}} & 14 & 20 & 37 & 63 & 87 & \textbf{70.7\%} & \textbf{89.5\%}\\
          \textbf{PKD} & \textbf{44.1} & \textbf{50.4} & \textbf{53.5} & \textbf{55.5} & \textbf{76.1} & \textbf{76.6\%}& \textbf{89.5\%} \\
          \bottomrule
        \end{tabular}
      }
    }
\end{table}

% \end{document}

% \documentclass{article}
% \usepackage{multirow}
% \usepackage{booktabs}
% \usepackage{graphicx} % for \resizebox

% \begin{document}

\begin{table}[t]
    \centering
    \caption{Computation Cost in the Dirichlet Partition Scenario (Unit: Same as Table~9)
}\label{tab_tex:10}  
    \makebox[\linewidth][c]{
      \resizebox{1.0\linewidth}{!}{
        \begin{tabular}{c ccccc cc}   % 7c
          \toprule
          \textbf{Setup} & \multicolumn{7}{c}{\textbf{FashionMNIST, Dir(0.5), \#client=20}} \\
          \midrule
          \multirow{2}{*}{\textbf{Method}} & \multicolumn{5}{c}{\textbf{Target Min. Acc.}}& \multicolumn{2}{c}{\textbf{Final Acc.}} \\
          \cmidrule{2-6} \cmidrule(lr){7-8}
           & 30\% & 45\% & 50\% & 55\% & 60\% &   \textbf{Min.} & \textbf{Ave.}\\
          \midrule
          \textbf{FedAvg~\cite{fedavg}} & 16 & 27 & 41 & - & - & \textbf{53.1\%} & \textbf{86.0\%} \\
          \textbf{FedProx~\cite{FedProx}} & 18 & 25 & 36 & 48 & - & \textbf{56.9\%}  & \textbf{86.2\%} \\
          \textbf{FedCodl~\cite{FedCodl}} & 20 & 28 & 35 & 46 & - & \textbf{55.1\%}  & \textbf{86.6\%} \\
          \textbf{PKD} & \textbf{25.2} & \textbf{30.4} & \textbf{35.5} & \textbf{43.6} & \textbf{58.9} & \textbf{63.2\%} & \textbf{86.2\%}\\
          \bottomrule
        \end{tabular}
      }
    }
\end{table}

% \end{document}

%% Figure 11: 
\begin{figure}[!t]
  \centering
      \includegraphics[width=0.485\textwidth]{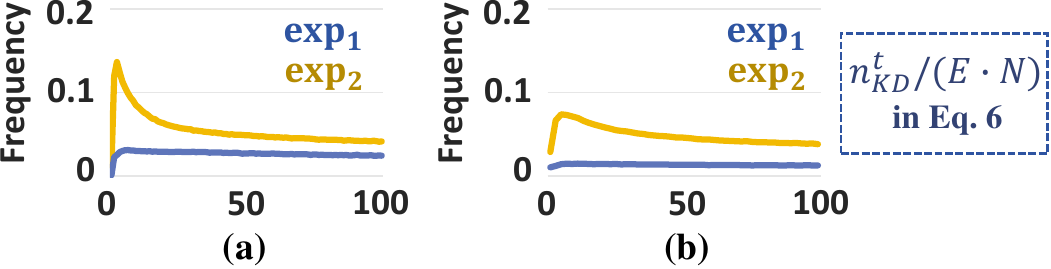} 	
      \caption{Expert Activation Frequency for FashionMNIST under (a) the Local Class-Balanced Scenario and (b) the Dirichlet Partition $Dir(0.5)$ Scenario.    
      }
      \label{fig: 11}
\end{figure}

In the proposed PKD method, the computational overhead mainly comes from expert training at Stage 2 and the computation of soft labels ($p_e$) during Stage 3. 
As shown in Eq.~\ref{eq: 6}, the computation cost of expert training is proportional to the total sample size of the weak classes ($\sum_{i} n_{g_i}$).
The actual computation cost for expert training ranges from 2.1$U$ to 40$U$ under different setups, where $U$ is the computation load of a local training round in the baseline scenario (FedAvg), as described in Eq.~\ref{eq: 4}. 
For Stage 3, the computational overhead primarily depends on the value of $\frac{n_{KD}^t}{E \cdot N}$ in Eq.~\ref{eq: 7}. 
Fig.~\ref{fig: 11} shows the frequency with which an expert is activated for KD. The value of $\frac{n_{KD}^t}{E \cdot N}$ increases rapidly at the onset of training and then gradually decreases as training progresses. KD occurs less frequently because, as the performance of the student model improves during training, fewer samples from the weak classes are misclassified.
Since PKD is applied only under specific conditions, the computational overhead for Stage 3 remains below 0.2$U$ per round across all benchmarks and setups in this work.

Table~\ref{tab_tex:9} reports the computation cost (measured in $U$) required to achieve the target minimum class-wise accuracy under the local class-balanced scenario. 
The symbol `–' indicates a failure to reach the target accuracy. 
The reported results for PKD reflect the total amount of computations across all three stages. Before the PKD stage, the model undergoes 20 rounds of warmup and 25 rounds of expert learning. The combined computational cost for these two phases is 32.5$U$. 
As a result, the overall computational overhead of PKD is relatively high initially, particularly when the minimum accuracy is below 50\%. 
Nonetheless, the model converges more rapidly than the baseline, requiring fewer computations to achieve the same level of accuracy in later training stages.
A similar trend has been observed under other setups. The computation cost in the Dirichlet partition scenario is shown in Table~\ref{tab_tex:10}. 
For FedCodl, we assume that the soft labels are precomputed and reused across multiple local epochs, and the inference cost of the teacher model is ignored. Besides, the computation cost for the proximal term in FedProx is also ignored for simplicity.

\vspace{-3pt}
\section{Conclusion}
\vspace{-2pt}
In this work, we investigate an interesting phenomenon that weak classes consistently exist even for class-balanced learning.
A class-specific partial knowledge distillation method is proposed to mitigate this inherent inter-class discrepancy.

\balance



\begin{thebibliography}{33}

%%% the Web address.

\ifx \showCODEN    \undefined \def \showCODEN     #1{\unskip}     \fi
\ifx \showISBNx    \undefined \def \showISBNx     #1{\unskip}     \fi
\ifx \showISBNxiii \undefined \def \showISBNxiii  #1{\unskip}     \fi
\ifx \showISSN     \undefined \def \showISSN      #1{\unskip}     \fi
\ifx \showLCCN     \undefined \def \showLCCN      #1{\unskip}     \fi
\ifx \shownote     \undefined \def \shownote      #1{#1}          \fi
\ifx \showarticletitle \undefined \def \showarticletitle #1{#1}   \fi
\ifx \showURL      \undefined \def \showURL       {\relax}        \fi
% invisible to TeX
\providecommand\bibfield[2]{#2}
\providecommand\bibinfo[2]{#2}
\providecommand\natexlab[1]{#1}
\providecommand\showeprint[2][]{arXiv:#2}

\bibitem[Buda et~al\mbox{.}(2018)]%
        {imbalance_cnn}
\bibfield{author}{\bibinfo{person}{Mateusz Buda}, \bibinfo{person}{Atsuto
  Maki}, {and} \bibinfo{person}{Maciej~A. Mazurowski}.}
  \bibinfo{year}{2018}\natexlab{}.
\newblock \showarticletitle{A systematic study of the class imbalance problem
  in convolutional neural networks}.
\newblock \bibinfo{journal}{\emph{Neural Networks}}  \bibinfo{volume}{106}
  (\bibinfo{year}{2018}), \bibinfo{pages}{249--259}.
\newblock
\showISSN{0893-6080}
\href{https://doi.org/10.1016/j.neunet.2018.07.011}{doi:\nolinkurl{10.1016/j.neunet.2018.07.011}}


\bibitem[Caldas et~al\mbox{.}(2018)]%
        {FEMNIST}
\bibfield{author}{\bibinfo{person}{Sebastian Caldas}, \bibinfo{person}{Peter
  Wu}, \bibinfo{person}{Tian Li}, \bibinfo{person}{Jakub Kone{\v{c}}n{\'y}},
  \bibinfo{person}{H.~Brendan McMahan}, \bibinfo{person}{Virginia Smith}, {and}
  \bibinfo{person}{Ameet Talwalkar}.} \bibinfo{year}{2018}\natexlab{}.
\newblock \showarticletitle{{LEAF:} {A} Benchmark for Federated Settings}.
\newblock \bibinfo{journal}{\emph{CoRR}}  \bibinfo{volume}{abs/1812.01097}
  (\bibinfo{year}{2018}).
\newblock
\showeprint[arXiv]{1812.01097}


\bibitem[Chawla et~al\mbox{.}(2002)]%
        {SMOTE}
\bibfield{author}{\bibinfo{person}{Nitesh~V. Chawla}, \bibinfo{person}{Kevin~W.
  Bowyer}, \bibinfo{person}{Lawrence~O. Hall}, {and} \bibinfo{person}{W.~Philip
  Kegelmeyer}.} \bibinfo{year}{2002}\natexlab{}.
\newblock \showarticletitle{{SMOTE:} Synthetic Minority Over-sampling
  Technique}.
\newblock \bibinfo{journal}{\emph{J. Artif. Intell. Res.}}
  \bibinfo{volume}{16} (\bibinfo{year}{2002}), \bibinfo{pages}{321--357}.
\newblock
\href{https://doi.org/10.1613/JAIR.953}{doi:\nolinkurl{10.1613/JAIR.953}}


\bibitem[Cui et~al\mbox{.}(2023)]%
        {ensemble}
\bibfield{author}{\bibinfo{person}{Jiequan Cui}, \bibinfo{person}{Shu Liu},
  \bibinfo{person}{Zhuotao Tian}, \bibinfo{person}{Zhisheng Zhong}, {and}
  \bibinfo{person}{Jiaya Jia}.} \bibinfo{year}{2023}\natexlab{}.
\newblock \showarticletitle{ResLT: Residual Learning for Long-Tailed
  Recognition}.
\newblock \bibinfo{journal}{\emph{IEEE Transactions on Pattern Analysis and
  Machine Intelligence}} \bibinfo{volume}{45}, \bibinfo{number}{3}
  (\bibinfo{year}{2023}), \bibinfo{pages}{3695--3706}.
\newblock
\href{https://doi.org/10.1109/TPAMI.2022.3174892}{doi:\nolinkurl{10.1109/TPAMI.2022.3174892}}


\bibitem[Cui et~al\mbox{.}(2019)]%
        {CBloss}
\bibfield{author}{\bibinfo{person}{Yin Cui}, \bibinfo{person}{Menglin Jia},
  \bibinfo{person}{Tsung{-}Yi Lin}, \bibinfo{person}{Yang Song}, {and}
  \bibinfo{person}{Serge~J. Belongie}.} \bibinfo{year}{2019}\natexlab{}.
\newblock \showarticletitle{Class-Balanced Loss Based on Effective Number of
  Samples}. In \bibinfo{booktitle}{\emph{{IEEE} Conference on Computer Vision
  and Pattern Recognition, {CVPR} 2019, Long Beach, CA, USA, June 16-20,
  2019}}. \bibinfo{publisher}{Computer Vision Foundation / {IEEE}},
  \bibinfo{pages}{9268--9277}.
\newblock
\href{https://doi.org/10.1109/CVPR.2019.00949}{doi:\nolinkurl{10.1109/CVPR.2019.00949}}


\bibitem[Devries and Taylor(2017)]%
        {cutout}
\bibfield{author}{\bibinfo{person}{Terrance Devries} {and}
  \bibinfo{person}{Graham~W. Taylor}.} \bibinfo{year}{2017}\natexlab{}.
\newblock \showarticletitle{Improved Regularization of Convolutional Neural
  Networks with Cutout}.
\newblock \bibinfo{journal}{\emph{CoRR}}  \bibinfo{volume}{abs/1708.04552}
  (\bibinfo{year}{2017}).
\newblock
\showeprint[arXiv]{1708.04552}
\urldef\tempurl%
\url{http://arxiv.org/abs/1708.04552}
\showURL{%
\tempurl}


\bibitem[Estabrooks et~al\mbox{.}(2004)]%
        {resample}
\bibfield{author}{\bibinfo{person}{Andrew Estabrooks}, \bibinfo{person}{Taeho
  Jo}, {and} \bibinfo{person}{Nathalie Japkowicz}.}
  \bibinfo{year}{2004}\natexlab{}.
\newblock \showarticletitle{A Multiple Resampling Method for Learning from
  Imbalanced Data Sets}.
\newblock \bibinfo{journal}{\emph{Comput. Intell.}} \bibinfo{volume}{20},
  \bibinfo{number}{1} (\bibinfo{year}{2004}), \bibinfo{pages}{18--36}.
\newblock
\href{https://doi.org/10.1111/J.0824-7935.2004.T01-1-00228.X}{doi:\nolinkurl{10.1111/J.0824-7935.2004.T01-1-00228.X}}


\bibitem[He and Garcia(2009)]%
        {imbalance_survey}
\bibfield{author}{\bibinfo{person}{Haibo He} {and} \bibinfo{person}{Edwardo~A.
  Garcia}.} \bibinfo{year}{2009}\natexlab{}.
\newblock \showarticletitle{Learning from Imbalanced Data}.
\newblock \bibinfo{journal}{\emph{IEEE Transactions on Knowledge and Data
  Engineering}} \bibinfo{volume}{21}, \bibinfo{number}{9}
  (\bibinfo{year}{2009}), \bibinfo{pages}{1263--1284}.
\newblock
\href{https://doi.org/10.1109/TKDE.2008.239}{doi:\nolinkurl{10.1109/TKDE.2008.239}}


\bibitem[He et~al\mbox{.}(2016)]%
        {ResNet}
\bibfield{author}{\bibinfo{person}{Kaiming He}, \bibinfo{person}{Xiangyu
  Zhang}, \bibinfo{person}{Shaoqing Ren}, {and} \bibinfo{person}{Jian Sun}.}
  \bibinfo{year}{2016}\natexlab{}.
\newblock \showarticletitle{Deep Residual Learning for Image Recognition}. In
  \bibinfo{booktitle}{\emph{2016 IEEE Conference on Computer Vision and Pattern
  Recognition (CVPR)}}. \bibinfo{pages}{770--778}.
\newblock
\href{https://doi.org/10.1109/CVPR.2016.90}{doi:\nolinkurl{10.1109/CVPR.2016.90}}


\bibitem[Hinton et~al\mbox{.}(2015)]%
        {KLD}
\bibfield{author}{\bibinfo{person}{Geoffrey Hinton}, \bibinfo{person}{Oriol
  Vinyals}, {and} \bibinfo{person}{Jeff Dean}.}
  \bibinfo{year}{2015}\natexlab{}.
\newblock \bibinfo{title}{Distilling the Knowledge in a Neural Network}.
\newblock
\showeprint[arxiv]{1503.02531}~[stat.ML]
\urldef\tempurl%
\url{https://arxiv.org/abs/1503.02531}
\showURL{%
\tempurl}


\bibitem[Kim et~al\mbox{.}(2020)]%
        {m2m}
\bibfield{author}{\bibinfo{person}{Jaehyung Kim}, \bibinfo{person}{Jongheon
  Jeong}, {and} \bibinfo{person}{Jinwoo Shin}.}
  \bibinfo{year}{2020}\natexlab{}.
\newblock \showarticletitle{M2m: Imbalanced Classification via Major-to-Minor
  Translation}. In \bibinfo{booktitle}{\emph{2020 {IEEE/CVF} Conference on
  Computer Vision and Pattern Recognition, {CVPR} 2020, Seattle, WA, USA, June
  13-19, 2020}}. \bibinfo{publisher}{Computer Vision Foundation / {IEEE}},
  \bibinfo{pages}{13893--13902}.
\newblock
\href{https://doi.org/10.1109/CVPR42600.2020.01391}{doi:\nolinkurl{10.1109/CVPR42600.2020.01391}}


\bibitem[Krizhevsky(2009)]%
        {Cifar10}
\bibfield{author}{\bibinfo{person}{Alex Krizhevsky}.}
  \bibinfo{year}{2009}\natexlab{}.
\newblock \bibinfo{booktitle}{\emph{Learning multiple layers of features from
  tiny images}}.
\newblock \bibinfo{type}{{T}echnical {R}eport}. \bibinfo{institution}{Univ. of
  Toronto}, \bibinfo{address}{Toronto, Canada}.
\newblock


\bibitem[Lecun et~al\mbox{.}(1998)]%
        {MNIST}
\bibfield{author}{\bibinfo{person}{Y. Lecun}, \bibinfo{person}{L. Bottou},
  \bibinfo{person}{Y. Bengio}, {and} \bibinfo{person}{P. Haffner}.}
  \bibinfo{year}{1998}\natexlab{}.
\newblock \showarticletitle{Gradient-based learning applied to document
  recognition}.
\newblock \bibinfo{journal}{\emph{Proc. IEEE}} \bibinfo{volume}{86},
  \bibinfo{number}{11} (\bibinfo{year}{1998}), \bibinfo{pages}{2278--2324}.
\newblock
\href{https://doi.org/10.1109/5.726791}{doi:\nolinkurl{10.1109/5.726791}}


\bibitem[Li et~al\mbox{.}(2023)]%
        {WeightAggregation}
\bibfield{author}{\bibinfo{person}{Zexi Li}, \bibinfo{person}{Tao Lin},
  \bibinfo{person}{Xinyi Shang}, {and} \bibinfo{person}{Chao Wu}.}
  \bibinfo{year}{2023}\natexlab{}.
\newblock \showarticletitle{Revisiting weighted aggregation in federated
  learning with neural networks}. In \bibinfo{booktitle}{\emph{Proceedings of
  the 40th International Conference on Machine Learning}} (Honolulu, Hawaii,
  USA) \emph{(\bibinfo{series}{ICML'23})}. \bibinfo{publisher}{JMLR.org},
  Article \bibinfo{articleno}{816}, \bibinfo{numpages}{22}~pages.
\newblock


\bibitem[Li et~al\mbox{.}(2022)]%
        {extremeNonIID}
\bibfield{author}{\bibinfo{person}{Zijian Li}, \bibinfo{person}{Jiawei Shao},
  \bibinfo{person}{Yuyi Mao}, \bibinfo{person}{Jessie~Hui Wang}, {and}
  \bibinfo{person}{Jun Zhang}.} \bibinfo{year}{2022}\natexlab{}.
\newblock \showarticletitle{Federated learning with gan-based data synthesis
  for non-iid clients}. In \bibinfo{booktitle}{\emph{International workshop on
  trustworthy federated learning}}. Springer, \bibinfo{pages}{17--32}.
\newblock


\bibitem[Lin et~al\mbox{.}(2020)]%
        {Focal}
\bibfield{author}{\bibinfo{person}{Tsung{-}Yi Lin}, \bibinfo{person}{Priya
  Goyal}, \bibinfo{person}{Ross~B. Girshick}, \bibinfo{person}{Kaiming He},
  {and} \bibinfo{person}{Piotr Doll{\'{a}}r}.} \bibinfo{year}{2020}\natexlab{}.
\newblock \showarticletitle{Focal Loss for Dense Object Detection}.
\newblock \bibinfo{journal}{\emph{{IEEE} Trans. Pattern Anal. Mach. Intell.}}
  \bibinfo{volume}{42}, \bibinfo{number}{2} (\bibinfo{year}{2020}),
  \bibinfo{pages}{318--327}.
\newblock
\href{https://doi.org/10.1109/TPAMI.2018.2858826}{doi:\nolinkurl{10.1109/TPAMI.2018.2858826}}


\bibitem[Liu et~al\mbox{.}(2022)]%
        {EMANATE}
\bibfield{author}{\bibinfo{person}{Bo Liu}, \bibinfo{person}{Haoxiang Li},
  \bibinfo{person}{Hao Kang}, \bibinfo{person}{Gang Hua}, {and}
  \bibinfo{person}{Nuno Vasconcelos}.} \bibinfo{year}{2022}\natexlab{}.
\newblock \showarticletitle{Breadcrumbs: Adversarial Class-Balanced Sampling
  for Long-Tailed Recognition}. In \bibinfo{booktitle}{\emph{Computer Vision --
  ECCV 2022}}, \bibfield{editor}{\bibinfo{person}{Shai Avidan},
  \bibinfo{person}{Gabriel Brostow}, \bibinfo{person}{Moustapha Ciss{\'e}},
  \bibinfo{person}{Giovanni~Maria Farinella}, {and} \bibinfo{person}{Tal
  Hassner}} (Eds.). \bibinfo{publisher}{Springer Nature Switzerland},
  \bibinfo{address}{Cham}, \bibinfo{pages}{637--653}.
\newblock
\showISBNx{978-3-031-20053-3}


\bibitem[Liu et~al\mbox{.}(2020)]%
        {LEAP}
\bibfield{author}{\bibinfo{person}{Jialun Liu}, \bibinfo{person}{Yifan Sun},
  \bibinfo{person}{Chuchu Han}, \bibinfo{person}{Zhaopeng Dou}, {and}
  \bibinfo{person}{Wenhui Li}.} \bibinfo{year}{2020}\natexlab{}.
\newblock \showarticletitle{Deep Representation Learning on Long-Tailed Data:
  {A} Learnable Embedding Augmentation Perspective}. In
  \bibinfo{booktitle}{\emph{2020 {IEEE/CVF} Conference on Computer Vision and
  Pattern Recognition, {CVPR} 2020, Seattle, WA, USA, June 13-19, 2020}}.
  \bibinfo{publisher}{Computer Vision Foundation / {IEEE}},
  \bibinfo{pages}{2967--2976}.
\newblock
\href{https://doi.org/10.1109/CVPR42600.2020.00304}{doi:\nolinkurl{10.1109/CVPR42600.2020.00304}}


\bibitem[McMahan et~al\mbox{.}(2017)]%
        {fedavg}
\bibfield{author}{\bibinfo{person}{Brendan McMahan}, \bibinfo{person}{Eider
  Moore}, \bibinfo{person}{Daniel Ramage}, \bibinfo{person}{Seth Hampson},
  {and} \bibinfo{person}{Blaise Aguera~y Arcas}.}
  \bibinfo{year}{2017}\natexlab{}.
\newblock \showarticletitle{{Communication-Efficient Learning of Deep Networks
  from Decentralized Data}}. In \bibinfo{booktitle}{\emph{Proceedings of the
  20th International Conference on Artificial Intelligence and Statistics}}
  \emph{(\bibinfo{series}{Proceedings of Machine Learning Research},
  Vol.~\bibinfo{volume}{54})}, \bibfield{editor}{\bibinfo{person}{Aarti Singh}
  {and} \bibinfo{person}{Jerry Zhu}} (Eds.). \bibinfo{publisher}{PMLR},
  \bibinfo{pages}{1273--1282}.
\newblock
\urldef\tempurl%
\url{https://proceedings.mlr.press/v54/mcmahan17a.html}
\showURL{%
\tempurl}


\bibitem[Minka(2000)]%
        {DirichletDistribution}
\bibfield{author}{\bibinfo{person}{Thomas Minka}.}
  \bibinfo{year}{2000}\natexlab{}.
\newblock \bibinfo{title}{Estimating a Dirichlet distribution}.
\newblock


\bibitem[Ni et~al\mbox{.}(2022)]%
        {FedCodl}
\bibfield{author}{\bibinfo{person}{Xuanming Ni}, \bibinfo{person}{Xinyuan
  Shen}, {and} \bibinfo{person}{Huimin Zhao}.} \bibinfo{year}{2022}\natexlab{}.
\newblock \showarticletitle{Federated optimization via knowledge
  codistillation}.
\newblock \bibinfo{journal}{\emph{Expert Systems with Applications}}
  \bibinfo{volume}{191} (\bibinfo{year}{2022}), \bibinfo{pages}{116310}.
\newblock
\showISSN{0957-4174}
\href{https://doi.org/10.1016/j.eswa.2021.116310}{doi:\nolinkurl{10.1016/j.eswa.2021.116310}}


\bibitem[Ren et~al\mbox{.}(2020)]%
        {meta2}
\bibfield{author}{\bibinfo{person}{Jiawei Ren}, \bibinfo{person}{Cunjun Yu},
  \bibinfo{person}{shunan sheng}, \bibinfo{person}{Xiao Ma},
  \bibinfo{person}{Haiyu Zhao}, \bibinfo{person}{Shuai Yi}, {and}
  \bibinfo{person}{hongsheng Li}.} \bibinfo{year}{2020}\natexlab{}.
\newblock \showarticletitle{Balanced Meta-Softmax for Long-Tailed Visual
  Recognition}. In \bibinfo{booktitle}{\emph{Advances in Neural Information
  Processing Systems}}, \bibfield{editor}{\bibinfo{person}{H.~Larochelle},
  \bibinfo{person}{M.~Ranzato}, \bibinfo{person}{R.~Hadsell},
  \bibinfo{person}{M.F. Balcan}, {and} \bibinfo{person}{H.~Lin}} (Eds.),
  Vol.~\bibinfo{volume}{33}. \bibinfo{publisher}{Curran Associates, Inc.},
  \bibinfo{pages}{4175--4186}.
\newblock


\bibitem[Sahu et~al\mbox{.}(2018)]%
        {FedProx}
\bibfield{author}{\bibinfo{person}{Anit~Kumar Sahu}, \bibinfo{person}{Tian Li},
  \bibinfo{person}{Maziar Sanjabi}, \bibinfo{person}{Manzil Zaheer},
  \bibinfo{person}{Ameet Talwalkar}, {and} \bibinfo{person}{Virginia Smith}.}
  \bibinfo{year}{2018}\natexlab{}.
\newblock \showarticletitle{On the Convergence of Federated Optimization in
  Heterogeneous Networks}.
\newblock \bibinfo{journal}{\emph{CoRR}}  \bibinfo{volume}{abs/1812.06127}
  (\bibinfo{year}{2018}).
\newblock
\showeprint[arXiv]{1812.06127}


\bibitem[Shuai et~al\mbox{.}(2022)]%
        {BalanceFL}
\bibfield{author}{\bibinfo{person}{Xian Shuai}, \bibinfo{person}{Yulin Shen},
  \bibinfo{person}{Siyang Jiang}, \bibinfo{person}{Zhihe Zhao},
  \bibinfo{person}{Zhenyu Yan}, {and} \bibinfo{person}{Guoliang Xing}.}
  \bibinfo{year}{2022}\natexlab{}.
\newblock \showarticletitle{BalanceFL: Addressing Class Imbalance in Long-Tail
  Federated Learning}. In \bibinfo{booktitle}{\emph{21st {ACM/IEEE}
  International Conference on Information Processing in Sensor Networks, {IPSN}
  2022, Milano, Italy, May 4-6, 2022}}. \bibinfo{publisher}{{IEEE}},
  \bibinfo{pages}{271--284}.
\newblock
\href{https://doi.org/10.1109/IPSN54338.2022.00029}{doi:\nolinkurl{10.1109/IPSN54338.2022.00029}}


\bibitem[Simonyan and Zisserman(2014)]%
        {vgg}
\bibfield{author}{\bibinfo{person}{Karen Simonyan} {and}
  \bibinfo{person}{Andrew Zisserman}.} \bibinfo{year}{2014}\natexlab{}.
\newblock \showarticletitle{{Very deep convolutional networks for large-scale
  image recognition}}.
\newblock \bibinfo{journal}{\emph{arXiv preprint arXiv:1409.1556}}
  (\bibinfo{year}{2014}).
\newblock


\bibitem[Wang et~al\mbox{.}(2020)]%
        {FedMA}
\bibfield{author}{\bibinfo{person}{Hongyi Wang}, \bibinfo{person}{Mikhail
  Yurochkin}, \bibinfo{person}{Yuekai Sun}, \bibinfo{person}{Dimitris
  Papailiopoulos}, {and} \bibinfo{person}{Yasaman Khazaeni}.}
  \bibinfo{year}{2020}\natexlab{}.
\newblock \bibinfo{title}{Federated Learning with Matched Averaging}.
\newblock
\showeprint[arxiv]{2002.06440}~[cs.LG]
\urldef\tempurl%
\url{https://arxiv.org/abs/2002.06440}
\showURL{%
\tempurl}


\bibitem[Xiang et~al\mbox{.}(2020)]%
        {LFME}
\bibfield{author}{\bibinfo{person}{Liuyu Xiang}, \bibinfo{person}{Guiguang
  Ding}, {and} \bibinfo{person}{Jungong Han}.} \bibinfo{year}{2020}\natexlab{}.
\newblock \showarticletitle{Learning From Multiple Experts: Self-paced
  Knowledge Distillation for Long-Tailed Classification}. In
  \bibinfo{booktitle}{\emph{Computer Vision - {ECCV} 2020 - 16th European
  Conference, Glasgow, UK, August 23-28, 2020, Proceedings, Part {V}}}
  \emph{(\bibinfo{series}{Lecture Notes in Computer Science},
  Vol.~\bibinfo{volume}{12350})}, \bibfield{editor}{\bibinfo{person}{Andrea
  Vedaldi}, \bibinfo{person}{Horst Bischof}, \bibinfo{person}{Thomas Brox},
  {and} \bibinfo{person}{Jan{-}Michael Frahm}} (Eds.).
  \bibinfo{publisher}{Springer}, \bibinfo{pages}{247--263}.
\newblock
\href{https://doi.org/10.1007/978-3-030-58558-7\_15}{doi:\nolinkurl{10.1007/978-3-030-58558-7\_15}}


\bibitem[Xiao et~al\mbox{.}(2017)]%
        {Fashion}
\bibfield{author}{\bibinfo{person}{Han Xiao}, \bibinfo{person}{Kashif Rasul},
  {and} \bibinfo{person}{Roland Vollgraf}.} \bibinfo{year}{2017}\natexlab{}.
\newblock \showarticletitle{Fashion-MNIST: a Novel Image Dataset for
  Benchmarking Machine Learning Algorithms}.
\newblock \bibinfo{journal}{\emph{CoRR}}  \bibinfo{volume}{abs/1708.07747}
  (\bibinfo{year}{2017}).
\newblock
\showeprint[arXiv]{1708.07747}


\bibitem[Yao et~al\mbox{.}(2023)]%
        {KDrelevant2}
\bibfield{author}{\bibinfo{person}{Dezhong Yao}, \bibinfo{person}{Wanning Pan},
  \bibinfo{person}{Yutong Dai}, \bibinfo{person}{Yao Wan},
  \bibinfo{person}{Xiaofeng Ding}, \bibinfo{person}{Chen Yu},
  \bibinfo{person}{Hai Jin}, \bibinfo{person}{Zheng Xu}, {and}
  \bibinfo{person}{Lichao Sun}.} \bibinfo{year}{2023}\natexlab{}.
\newblock \showarticletitle{FedGKD: Toward heterogeneous federated learning via
  global knowledge distillation}.
\newblock \bibinfo{journal}{\emph{IEEE Trans. Comput.}} \bibinfo{volume}{73},
  \bibinfo{number}{1} (\bibinfo{year}{2023}), \bibinfo{pages}{3--17}.
\newblock


\bibitem[Yurochkin et~al\mbox{.}(2019)]%
        {DirichletFL}
\bibfield{author}{\bibinfo{person}{Mikhail Yurochkin}, \bibinfo{person}{Mayank
  Agarwal}, \bibinfo{person}{Soumya Ghosh}, \bibinfo{person}{Kristjan
  Greenewald}, \bibinfo{person}{Nghia Hoang}, {and} \bibinfo{person}{Yasaman
  Khazaeni}.} \bibinfo{year}{2019}\natexlab{}.
\newblock \showarticletitle{Bayesian nonparametric federated learning of neural
  networks}. In \bibinfo{booktitle}{\emph{International conference on machine
  learning}}. PMLR, \bibinfo{pages}{7252--7261}.
\newblock


\bibitem[Zhang et~al\mbox{.}(2023b)]%
        {surveyCIFL}
\bibfield{author}{\bibinfo{person}{Jing Zhang}, \bibinfo{person}{Chuanwen Li},
  \bibinfo{person}{Jianzgong Qi}, {and} \bibinfo{person}{Jiayuan He}.}
  \bibinfo{year}{2023}\natexlab{b}.
\newblock \bibinfo{title}{A Survey on Class Imbalance in Federated Learning}.
\newblock
\showeprint[arxiv]{2303.11673}~[cs.LG]
\urldef\tempurl%
\url{https://arxiv.org/abs/2303.11673}
\showURL{%
\tempurl}


\bibitem[Zhang et~al\mbox{.}(2023a)]%
        {surveyLT}
\bibfield{author}{\bibinfo{person}{Yifan Zhang}, \bibinfo{person}{Bingyi Kang},
  \bibinfo{person}{Bryan Hooi}, \bibinfo{person}{Shuicheng Yan}, {and}
  \bibinfo{person}{Jiashi Feng}.} \bibinfo{year}{2023}\natexlab{a}.
\newblock \showarticletitle{Deep Long-Tailed Learning: {A} Survey}.
\newblock \bibinfo{journal}{\emph{{IEEE} Trans. Pattern Anal. Mach. Intell.}}
  \bibinfo{volume}{45}, \bibinfo{number}{9} (\bibinfo{year}{2023}),
  \bibinfo{pages}{10795--10816}.
\newblock
\href{https://doi.org/10.1109/TPAMI.2023.3268118}{doi:\nolinkurl{10.1109/TPAMI.2023.3268118}}


\bibitem[Zhao et~al\mbox{.}(2020)]%
        {clientsampl}
\bibfield{author}{\bibinfo{person}{Fengpan Zhao}, \bibinfo{person}{Yan Huang},
  \bibinfo{person}{Akshita Maradapu Vera~Venkata Sai}, {and}
  \bibinfo{person}{Yubao Wu}.} \bibinfo{year}{2020}\natexlab{}.
\newblock \showarticletitle{A Cluster-based Solution to Achieve Fairness in
  Federated Learning}. In \bibinfo{booktitle}{\emph{2020 IEEE Intl Conf on
  Parallel \& Distributed Processing with Applications, Big Data \& Cloud
  Computing, Sustainable Computing \& Communications, Social Computing \&
  Networking (ISPA/BDCloud/SocialCom/SustainCom)}}. \bibinfo{pages}{875--882}.
\newblock
\href{https://doi.org/10.1109/ISPA-BDCloud-SocialCom-SustainCom51426.2020.00135}{doi:\nolinkurl{10.1109/ISPA-BDCloud-SocialCom-SustainCom51426.2020.00135}}


\end{thebibliography}
\end{document}